\documentclass[a4paper, twocolumn]{article}  
\pdfoutput=1
\usepackage[utf8]{inputenc}
\usepackage[english]{babel}
\usepackage[left=2cm,right=2cm,top=2cm,bottom=2cm]{geometry}
\usepackage{amsmath}
\usepackage{amsfonts}
\usepackage{amssymb}
\usepackage{booktabs}
\usepackage{harvard}
\usepackage{pdfpages}

\usepackage[linesnumbered,ruled]{algorithm2e}
\usepackage[noend]{algpseudocode}
\usepackage{indentfirst}

\usepackage{sansmath}
\sansmath 

\usepackage[colorlinks]{hyperref}
\hypersetup{
	allcolors=black
}

\title{Towards a Common Implementation of Reinforcement Learning for Multiple Robotic Tasks}

\author{Angel Mart\'inez-Tenor, Juan Antonio Fern\'andez-Madrigal, Ana Cruz-Mart\'in and \\ Javier  Gonz\'alez-Jim\'enez \\
\\ 
\small{Machine Perception and Intelligent Robotics (MAPIR). Dept. Ingenier\'ia de Sistemas y Autom\'atica} \\ 
\small{Instituto de Investigaci\'on Biom\'edica de M\'alaga, Universidad de M\'alaga. Spain.}}

\date{}

\begin{document}

\twocolumn[
\begin{@twocolumnfalse} 
	\maketitle

	\begin{abstract}
		Mobile robots are increasingly being employed for performing complex tasks in dynamic environments. Reinforcement learning (RL) methods are recognized to be promising for specifying such tasks in a relatively simple manner. However, the strong dependency between the learning method and the task to learn is a well-known problem that restricts practical implementations of RL in robotics, often requiring major modifications of parameters and adding other techniques for each particular task. In this paper we present a practical core implementation of RL which enables the learning process for multiple robotic tasks with minimal per-task tuning or none. Based on value iteration methods, this implementation includes a novel approach for action selection, called Q-biased softmax regression (QBIASSR), which avoids poor performance of the learning process when the robot reaches new unexplored states. Our approach takes advantage of the structure of the state space by attending the physical variables involved (e.g., distances to obstacles, $X,Y,\theta$ pose, etc.), thus experienced sets of states may favor the decision-making process of unexplored or rarely-explored states. This improvement has a relevant role in reducing the tuning of the algorithm for particular tasks. Experiments with real and simulated robots, performed with the software framework also introduced here, show that our implementation is effectively able to learn different robotic tasks without tuning the learning method. Results also suggest that the combination of true online SARSA($\lambda$) (TOSL) with QBIASSR can outperform the existing RL core algorithms in low-dimensional robotic tasks.
	\end{abstract}\vspace*{1cm}
\end{@twocolumnfalse}
]

\section{Introduction}\label{sec:intro}

The operation of mobile robots in real environments requires coping with unforeseen situations that human programmers find difficult or impossible to catalog and therefore to code into a program. Therefore, it is highly desirable that the robot itself modifies its behavior to cope with them. An ideal solution would consist of an architecture capable of automatically evolving the controller to improve the performance of any task over time; the same implementation should be employed in different robots and environments without considering the scenario or the hardware involved, and without reprogramming or manually tuning the resulting controllers, i.e., such architecture could be employed for new tasks by making minor changes to already implemented ones, e.g., by reprogramming the objective and the devices required to accomplish them. 

We can consider two paradigmatic examples of robotics projects that would be benefited from such an ideal, found in our research group, but common in many others: i) The implementation of several pick and place tasks with a complex 6DOF low-weight robotic arm, whose  model is not available but that is required to execute complicated trajectories; in this case, top-down implementations demanded many hours of work, usually employed to adapt small variations of the same task. ii) A set of mobile robotics platforms that must be used for autonomous navigation, teleoperation, and collaborative control; differences between the behavior of the mobile bases, even among identical robots, often require the navigation programs to be tuned from one robot to another, and other features such as the variations of the wear of the wheels and the battery levels contribute to worsen the performance of the tasks as well.

The work presented in this paper is a first step towards the ideal solution explained before. It focuses on building a common implementation of a machine learning core approach intended for solving multiple low-dimensional tasks found in service robotics, such as wandering, 2D mobile navigation, 3D arm motion, etc. For this purpose, the robots should learn by themselves the effects of their actions, with minimal tuning by the engineer. A well-known machine learning and decision-making methodology, reinforcement learning (RL), has been chosen as the basic mechanism for that. RL is based on the explicit implementation, by the engineer, of a set of rewards that define the objectives of the task to learn.

Most RL publications, including those based on low-realistic simulated robots (see section \ref{sec:backandrelated}), show their advantages by measuring the eventual convergence to the optimal solution to the task at hand, or the level of performance, often based on the average obtained reward, after a relatively large number of steps, episodes, and repetitions of the learning process. Unfortunately, direct application of most of these methods in real scenarios would require days or months of learning to achieve a fair performance, even in relatively simple tasks such as 2D navigation in small environments. Consequently, other techniques must be added when dealing with RL problems in robotics to make them tractable, such as those based on a clever use of approximate representations, prior knowledge, and models. These methods, in turn, need to be carefully selected and hand-tuned for each specific task to learn, even for very closely related ones. All of this makes RL not applicable to robotics out of the box yet \cite{kober2013reinforcement}. 

The present work has been motivated for this lack of practical, task-independent RL mechanisms for real robots, and it contributes a core method suitable for learning multiple tasks, although possibly non-optimally. This core method intends to avoid the use of representations, prior knowledge and models, and to be extendable with most advanced RL techniques. We also present here a study of the performance of our core RL solution compared to combinations of basic RL algorithms, i.e., those updating their \mbox{$Q$-values} without using approximators and models, and also with several action selection techniques. We have performed empirical comparisons of the learning processes in different robotic tasks. At this state of our research, only the tabular case of RL with low-dimensional state spaces (easy hand-crafted feature representations) has been addressed, leaving generalization and function approximators, needed for higher-dimensional tasks, for future work. 

More concretely, our approach combines the recent true online SARSA($\lambda$) algorithm (TOSL)  \cite{van2014true} with a novel action selection technique. That combination outperforms the state-of-the-art RL methods for multiple low-dimensional robotic tasks. In particular, our action selection technique, called Q-BIASed Softmax Regression (QBIASSR), has a relevant role in reducing the effort of tuning the method for different tasks; it has been developed after observation of the learning processes in a realistic robotic simulator, especially when the robots reach new states that were closely related to already explored ones, something that occurs in many applications. QBIASSR copes with the information of those related states in a simple and intuitive way, using information from previous experiences to minimize exploration and therefore achieving good learning results independently of the particular task. Our contribution is completed with a new RL software framework for robotics called \mbox{RL-ROBOT}, developed from scratch with the intention of easing the inclusion of new tasks for robotics researchers. It is also useful as a hands-on tool for beginners. 

The structure of the paper is summarized as follows: section \ref{sec:backandrelated} provides a brief introduction to RL and the state of the art of its application to robotics. Our approach is described in section \ref{sec:approach}, that deals with both the combination of already existent techniques and the QBIASSR algorithm. Section \ref{sec:implementation} describes the \mbox{RL-ROBOT} software framework implemented for the simulated and real experiments  shown in section \ref{sec:experiments}. Finally, results and future work are summarized in section \ref{sec:conclusions}.

\section{Background and related work}\label{sec:backandrelated}
In this section we provide a brief summary of the core RL algorithms related to this work and their application to robotics. 

\subsection{Reinforcement learning}
The autonomous learning concept in robotics dates back to Alan Turing's idea of having robots that learn in a way similar to children \cite{turing1950computing}. This evolved to the modern concept of developmental robotics \cite{lungarella2003developmental}, a today still emergent paradigm analogue to the one of mental development in human psychology. In autonomous learning, mechanisms for decision-making under uncertainty are an important and increasingly popular approach. They represent cognitive processes capable of selecting a sequence of actions that are expected to lead to a specific outcome. Markov decision processes (MDP) \cite{murphy2002dynamic} are the most exploited paradigm in this sense. They are dynamic Bayesian networks whose nodes define system states and whose arcs define actions along with their associated rewards and probability of occurrence. Formally, an MDP is a tuple $(S, A, T, R)$, where $S$ is a finite set of states, $A$ a finite set of actions, $T$ the state transition function, and $R$ the reward function, defined as \mbox{$R:(S\times A\times S)\rightarrow \mathbb{R}$}.  An important concept in MDPs is the one of policy, a function $ \pi :S \to A $ that defines the action $a$ to execute when the agent is in state $s$. The term policy value or $V$ indicates the goodness of a policy measured through some definition of the expected reward when it is executed. Most RL problems use the total expected discounted reward gathered after infinite steps of the decision process as the policy value, thus decreasing the importance of future rewards as the agent moves forward in time. The expected total reward from taking a specific action in a given state is defined then as:

\begin{align}
Q_\pi(s,a)& = \sum_{s'\in succ(s)}T(s,a,s')\cdot E[R(s,a,s')] \nonumber \\
& +\sum_{s'\in succ(s)}\gamma \cdot T(s,a,s') \cdot V_\pi (s')
\end{align}
being $\gamma \in (0, 1)$ the discount rate and $succ(s)$ the set of states that can be reached from state $s$. Then, $V(s) = max_a Q(s,a)$. 

With those definitions, Bellman equations \cite{bellman1956dynamic} can be used recursively for improving an arbitrary initial policy until it converges to the optimal one (the one with greatest value). A classical algorithm for finding optimal policies through Bellman equations is value iteration, which computes $Q_k(s,a)$ and $V_k(s) = max_a Q_k(s,a)$ incrementally in every step $k$. 

RL algorithms address the problem that arises in MDPs when no information about the transition function $T(s,a,s')$ is given, but can be estimated from the real system by making observations. In particular, \mbox{Q-learning}  \cite{watkins1989learning} is one of the most practical algorithms employed in RL; it belongs to the class of temporal difference (TD) learning methods, which combine several aspects of Monte Carlo and supervised programming for handling delayed rewards. Equation \ref{eq:qlearning} represents the general form of \mbox{Q-learning}, used in this work as the algorithm of reference: 
\begin{align}
Q_k(s,a) = Q_{k-1}(s,a)+\alpha_k\cdot\delta_k
\label{eq:qlearning}
\end{align}
where $\alpha_k$ is the learning rate  $\in (0,1)$, and $\delta_k$ the TD error, defined as:
\begin{align}
\delta_k = R + \gamma \cdot V_{k-1}(s')-Q_{k-1}(s,a)  
\end{align}

\begin{algorithm}
	\For{every learning step k} {
		select new action $a$ (exploration strategy) \\
		execute $a$ (wait step-time if needed)\\
		observe reached state $s'$ and get reward $R$ \\
		update $Q(s,a)$:
		\, $\delta_k \leftarrow R + \gamma \cdot V_{k-1}(s')-Q_{k-1}(s,a)$
		\, $Q_k(s,a) \leftarrow Q_{k-1}(s,a)+\alpha_k\cdot\delta_k$ \\
		$V_k(s) = max_a Q_k(s,a)$ \\
		$s \leftarrow s'$
	}
	\caption{Q-learning}
	\label{alg:qlearning}
\end{algorithm}

The success of \mbox{Q-learning} (in pseudocode in algorithm \ref{alg:qlearning}), as well as of most RL methods, depends on the accurate choice of the parameters $\alpha$ and $\gamma$, along with a set of suitable rewards $R(s,a,s')$, that define the task to learn, and an action selection strategy. The latter refers us to the exploitation-exploration dilemma, which consists in deciding  whether the agent should exploit its current learned policy or explore other actions at each learning step.  

A practical upgrade of \mbox{Q-learning} came with SARSA \cite{rummery1994line}, which updates the \mbox{$Q$-values} for the action the agent will execute in the next learning step instead of the optimal action learned so far. Hence $a'$, the action to perform in the next step, must also be selected before updating $Q(s,a)$. The TD error in this case is:
\begin{align}
\delta_k = R +  \gamma \cdot Q_{k-1}(s',a')-Q_{k-1}(s,a)	
\end{align}

SARSA often reduces the number of steps needed for learning a task, and it also avoids policies close to large negative rewards, which could harm a real robot.

Another mechanism widely used for improving RL is eligibility traces (ET) \cite{klopf1972brain}, \cite{kaelbling1996reinforcement}, \cite{sutton1998reinforcement} and \cite{lin1992reinforcement}, and its variant replacing traces \cite{singh1996reinforcement}. Based on a time decaying memory trace, this mechanism is able to simultaneously update many \mbox{$Q$-values} in each step, accelerating the learning process: when an agent using ET reaches a state associated with a large reward, the previously visited \mbox{$Q$-values} will receive a decaying fraction of this reward, the more recent the higher the reward; this encourages the execution of high-rewarded sequences of actions in future decisions. On-policy RL algorithms such as SARSA are directly compatible with ET, resulting in the mechanistic or backward view of TD($\lambda$) \cite{sutton1998reinforcement}; The ET factor $\lambda \in (0,1)$ must be accurately set to define the fraction of the reward to feedback. A recent improvement to TD($\lambda$) came with true online TD learning \cite{van2014true}, which addresses the bias-variance dilemma of the former. Particularly, the true online SARSA($\lambda$) algorithm (TOSL) has demonstrated having both better theoretical properties and empirically more efficient learning processes than traditional TD($\lambda$) algorithms \cite{van2015true}. 

So far the algorithms mentioned are derived from the classical value iteration, but there are other effective TD-based algorithms, such as actor-critic methods \cite{barto1983neuronlike}, R-learning \cite{schwartz1993reinforcement}, and those based on policy iteration. We found them less suitable as practical core RL methods for multiple tasks since they explicitly bias the learning process by adding detailed knowledge of the task. For example, policy iteration is often tuned to avoid extensive searches in the policy space in order to guarantee a good performance of the learning process. Other contributions that also fall out of the scope of this work include the combinations of RL algorithms with other mechanisms and function approximators that help with high-dimensional tasks, such as those from machine learning (e.g., Monte Carlo methods, dynamic programming, and exhaustive search), hierarchical RL, batch RL, neural RL, or that extend the problem, such as partially observable Markov decision processes (POMDP), policy gradient RL, inverse RL, bioinspired RL, etc. \cite{kaelbling1996reinforcement} and \cite{wiering2012reinforcement}. Although in many works learning and exploiting a model of the environment have proved to boost the learning process \cite{Campbell2016} and \cite{sutton2012dyna}, like in prioritized sweeping \cite{moore1993prioritized}, \mbox{model-based} RL has not been included in this work so as to focus on the study of generic model-free algorithms, that nevertheless can be easily extended to \mbox{model-based} ones.  

A careful review of publications reveals the difficulty of RL convergence to quasi-optimal solutions still today: the learning process often results incomplete or falls into local minima when the algorithm parameters are not properly tuned. The problem worsens when trying to solve different tasks \cite{kaelbling1996reinforcement},  \cite{sutton1998reinforcement}, \cite{szepesvari2010algorithms} and \cite{wiering2012reinforcement}. An adequate approach to the exploration-exploitation dilemma \cite{thrun1992efficient}, \cite{wiering1999explorations}, \cite{tokic2011value}, \cite{lopes2012exploration}, \cite{hester2013learning} and \cite{pecka2014safe} is also proved to be essential for the performance of the learning process, especially for practical applications.

Finally, a noticeable improvement in RL have come with the recent works in deep reinforcement learning (DRL)  \cite{mnih2013playing}, \cite{lillicrap2015continuous}, \cite{mnih2016asynchronous}, \cite{stadie2015incentivizing}, \cite{schaul2015prioritized} and \cite{osband2016deep},  which successfully integrate the advantages of deep learning with RL, leading to effective learning processes for high-dimensional tasks. Integrating such approaches in our core algorithm has been planned for future work for the reasons commented already.

\subsection{Reinforcement learning in robotics}
Today the efficacy of RL algorithms has been demonstrated in many different fields, such as game theory, control engineering, statistics, or even robotics when toy models or very low-realistic robot simulators are used \cite{sutton1998reinforcement}, \cite{kaelbling1996reinforcement}, \cite{wiering2012reinforcement} and \cite{kober2013reinforcement}. RL researchers can also resort to more advanced techniques to deal with the limitations arising with the practical aspects of RL, like the curse of dimensionality \cite{bellman1957}. 

However, when RL is applied to accurate physical realistic simulators or to real robots, we must face additional issues that are specific of this area, such as the curse of real-world samples \cite{kober2013reinforcement}, which obstruct the execution of RL-based methods, often making them unviable. Moreover, as RL in robotics is based on the environment-robot interaction, these difficulties are increased by the constraints posed by real-world time, often neglected in simulated scenarios where high number of steps and episodes are possible. Despite these limitations, many works have implemented RL-based methods for learning complex but particular robotic tasks, overcoming some of these problems \cite{lin1992reinforcement}, \cite{smart2002effective}, \cite{gaskett2002q}, \cite{abbeel2010autonomous}, \cite{yen2002reinforcement}, \cite{peters2003reinforcement}, \cite{ng2006autonomous}, \cite{wicaksono2011q}, \cite{degris2012model}, \cite{hester2012rtmba}, \cite{navarro2012real}, \cite{hester2013texplore}, \cite{kober2013reinforcement}, \cite{kormushev2013reinforcement}, \cite{vidal2013learning}, \cite{kober2014learning}, \cite{deisenroth2015gaussian} and \cite{garcia2015comprehensive}. At the time this work was in progress the first studies of DRL in robotics began to appear \cite{zhang2015towards} and \cite{sigaud2016towards}, addressing some high-dimensional tasks, usually with visual perception only.

The main conclusion to be drawn from this is that the complexity of RL in robotics requires the learning method to be complemented through advanced representation, prior knowledge, and models in order to make the problem tractable \cite{kober2013reinforcement}; this biases the learning process with knowledge from the task, resulting in a task-dependent learning method. Although some studies have searched for techniques adaptable to multiple tasks \cite{kober2014learning}, they usually rely on closely related ones, sometimes leading to poor performance.

\section{Reducing the task dependency: Overview of true online SARSA($\lambda$) with Q-biased softmax action selection} \label{sec:approach}

\begin{table*}[h!]
	\small\sf\centering
	\caption{RL core techniques evaluated in the preliminary tests}
	\begin{tabular}{p{0.28\textwidth}p{0.6\textwidth}}
		\toprule
		Core algorithms: & Q-learning, SARSA, SARSA($\lambda$), true online SARSA($\lambda$) \\
		\midrule
		Parameters sensibility: & learning rate $\alpha$, discount rate $\gamma$, ET factor $\lambda$, Boltzmann temperature $T$  \\
		\midrule
		Exploration-exploitation: & $\epsilon$-greedy, softmax regression (Boltzmann)\\
		\bottomrule
	\end{tabular}
	\label{core_techniques}
\end{table*}

An RL problem, defined as a set of states, actions, and rewards, looks abstract and task-independent by itself; however solving a task in a RL approach (i.e., converging to a near-optimal value function), depends on the structure of the unknown transition matrix, that is, on how easy is to explore all interesting states frequently enough. Therefore, our efforts to decrease the dependence on the particular task have been focused on accelerating the estimation of the \mbox{$Q$-values} for relevant states-actions pair that sometimes are difficult to explore in that task.

Firstly we have carried out a set of experiments to find a learning method among the existing RL techniques that has good properties when confronted with different tasks. The \mbox{RL-ROBOT} software framework, that will be described in depth in section \ref{sec:implementation}, has been used together with the very realistic \mbox{V-REP} robotic simulator \cite{vrep2013} to conduct these comparative experiments. The methods shown in table \ref{core_techniques} have been the ones evaluated for different robotic tasks, being the resulting average reward per step used to measure the performance of the learning processes for all of them. The setups and experiments of this preliminary stage will be described in more detail in sections \ref{sec:implementation} and \ref{sec:experiments} along with the rest of experiments.

These tests show that true online SARSA($\lambda$) (TOSL) is the best algorithm, with parameters $\lambda$ = 0.9, $\alpha$ = 0.1, and $\gamma$ = 0.9 that work well for different tasks; the exploration strategy with best results is softmax action selection with a Boltzmann distribution (equation \ref{eq:softmax}), with $T = 1$. 
\begin{align}
P(a) = \dfrac{e^{\dfrac{Q(s,a)}{T}}}{\sum_i e^{\dfrac{Q(s,a_i)}{T}}}
\label{eq:softmax}
\end{align}

Since eligibility traces (ET) update the value function in more states than the current one, therefore favoring the estimation of $V$ even when there are states that are difficult to explore frequently, we have enriched the method of the first experiments with a practical low-computational cost version of ET. The update rule of SARSA with ET, SARSA($\lambda$), is (compare with equation \ref{eq:qlearning}):
\begin{align}
Q_k(s,a)=Q_{k-1}(s,a)+\alpha\cdot\delta_k\cdot e_k(s,a)
\end{align}
where  $e_k(s,a)$ is the trace for $Q(s,a)$, being increased or set to 1 (replacing traces) whenever the state-action pair is explored, and decreased in every unexplored step as:
\begin{align}
e_k(s,a)=\gamma\cdot\lambda\cdot e_{k-1}(s,a) 
\label{eq:decaying_trace}
\end{align}
For a practical implementation, instead of updating all the $Q$-values at each learning step with their respective traces, a reduced ET register (RETR) has been defined that contains only a subset of traces with $e_k(s,a) \geq threshold$. We have empirically selected $threshold \geq 0.01$, i.e, those traces contributing less than 1\% of the current state-action pair (replacing traces case with $e  = 1$) were discarded. The number of steps $n$ elapsed from the last visit of a particular Q-value until its trace become negligible can be obtained from equation \ref{eq:decaying_trace} as:
\begin{align}
(\gamma\cdot\lambda)^n > threshold  \Rightarrow
n< \dfrac{\log(threshold)}{log(\gamma\cdot\lambda)}
\label{nlimit}
\end{align}
Applying the values resulting from the preliminary tests to equation \ref{nlimit} with $threshold = 0.01$ results in $n < 21.85$, i.e., RETR will be formed by the most recent 21 traces that will update $Q$ in each learning step, besides the current state. A comparative test between standard ET and reduced ET with $threshold = 0.01$ exhibits no differences in their learning performances, as shown in section \ref{sec:experiments}.

By observing the behavior of the robot on the simulator in these preliminary tests, we detected that the learning process was highly inefficient whenever the agent reached a new state with strong resemblance to already explored states; the robot had to learn from scratch an accurate action for that new state, or, in other words, intensive exploration was performed instead of trying suitable actions already learned in similar situations, resulting in an inaccurate and disappointing behavior from the human point of view. In order to avoid the use of task-dependent tuning for addressing this issue, we have devised the next contribution.

When designing a robotic task for a discrete state-space RL,  physical input variables involved in the learning process are usually identified, discretized, and hand-crafted before being mixed to generate structured states (structured states are thus states composed of a number of parts, each one representing an input variable \cite{moore1993prioritized}). Fortunately, by having identified these variables, areas of the $Q$ matrix associated to one or several combined set of them can be used for deciding what to do when an unexplored or poor-explored state is reached, i.e., we can use the information from states with similar values of these variables. This intuition has led us to the novel Q-biased softmax regression (QBIASSR).

In QBIASSR, when selecting an action to execute, instead of applying a softmax directly to the values of $Q(s)$ for the current state, this softmax is performed over another, virtual vector  $Q(s)_{biased}$, defined as:
\begin{align}
Q(s)_{biased} = Q(s) + bias(s)
\label{q_biased}
\end{align}
being $bias(s)$ a vector resulting from processing the values of other states of $Q$ that can be similar to $s$. This bias is calculated at each step of the learning process using averaged information from sets of states that share some structure with the current state $s$. In this way, the probabilities of choosing each action from $s$ will be biased by the previous experience obtained in similar states.

Many approaches can be used to define a suitable $bias(s)$, ranging from the processing of all the information of $Q$ prioritizing states with closer features to $s$, to use part of the state structure that has values similar to $s$; even being assisted by a hierarchy of sets of these parts. In our work we have defined $bias(s)$ empirically as the averaged contribution of subsets of values of $Q$. This turns out to be computationally simple and leads to the good results shown in section \ref{sec:experiments}. Each subset results from removing each input variable from the state space $S$ and averaging the \mbox{$Q$-values} for those states with equal values for the remaining input variables. A precise definition of QBIASSR is shown in algorithm \ref{alg:QBIASSR}.

\begin{algorithm}
	$A = \{a_1, a_2, ... a_p\}$ \Comment{the set of actions} \\
	$S = \{s_1, s_2, ... s_m\}$ \Comment{the set of states} \\
	$X = \{x_1, x_2, ... x_n\}$ \Comment{the vector of input variables} \\
	$X(s\in S) = \{x_1(s), x_2(s), ... x_n(s)\} $ \Comment {values of input variables for the state $s$} \\
	$SS(s\in S, x_i\in X) = $ subset of states $ ss \in S$ with $x_j(ss) = x_j(s) \quad \forall j\in [1,2,... n] \neq i$\\
	$SQ(s\in S, x_i\in X) = $ subset of $Q$ for $SS(s, x_i)$\\
	Agent in state $s$ must select and action $a'$ given $Q$  \\  
	\ForAll{$x_i \in x$} {
		$bias(s,x_i) \leftarrow$ avg($SQ(s,x_i)) \quad \forall ss \in SS(s,x_i)$         \Comment{resulting in a vector with same size of $Q(s)=[Q(s,a_1),Q(s,a_2),... Q(s,a_p)]$}
	}
	$bias(s) \leftarrow 1/n\cdot\sum_{i=0}^{n} bias(s,x_i)$ \\
	$Q(s)_{biased} \leftarrow Q(s) + bias(s)$   \\
	$a' \leftarrow $ softmax\_selection$(Q_{biased}, Temperature)$
	\caption{Q-biased softmax regression (QBIASSR)}
	\label{alg:QBIASSR}
\end{algorithm}

As an example, consider a simple mobile robotic agent with four discretized input variables $\{X, Y, \theta, d\}$, with $X \in \{0, 1\}$,  $Y \in  \{0, 1\}$, $\theta \in \{0, 90, -90\}$, and $d \in \{F, T\}$, thus composing 24 structured states. The agent has one output variable $MOV \in \{$advance, spin left, spin right$\}$ representing the available actions. If the robot reaches an unexplored state $s$ with $(X, Y, \theta, d) = (1, 1, 90, T)$ being $Q(s) = (0, 0, 0)$, a classical softmax will use an uniform distribution for selecting each available action since the state is unexplored. On the contrary, QBIASSR will create a biased distribution, involving those states with at least 3 values of $(X, Y, \theta, d)$ equal to $(1, 1, 90, T)$, as follows:
\begin{align*}
bias_x(s) &= \dfrac{Q_{(0,1,90, T)}+ Q_{(1,1,90,T)}}{2}\\
bias_y(s) &= \dfrac{Q_{(1,0,90, T)}+ Q_{(1,1,90, T)}}{2}\\
bias_{\theta}(s) &= \dfrac{Q_{(1,1,0, T)}+  Q_{(1,1,90, T)}+ Q_{(1,1,-90, T)}}{3}\\
bias_d(s) &= \dfrac{Q_{(1,1,90, T)}+ Q_{(1,1,90, F)}}{2}\\
bias(s) &=  \dfrac{bias_x(s)+ bias_y(s)+ bias_{\theta}(s) + bias_d(s)}{4}
\end{align*}

For assessing the goodness of our approach, several experimental comparatives between three settings, \mbox{Q-learning} with softmax regression (Q+SR), true online SARSA($\lambda$) with softmax regression (TOSL+SR), and true online SARSA($\lambda$) with Q-biased softmax regression (TOSL+QBIASSR), have been carried out. The experiments include wandering, 2D navigation, and 3D arm motion tasks, performed in sample-modeled simulations, V-REP simulations, and real-robot implementations. The setups and results will be described in sections \ref{sec:implementation} and \ref{sec:experiments}. This comparative study has proven that QBIASSR outperforms classical softmax action selection by improving the learning process. Neither convergence issues nor incompatibilities of this approach with approximation functions or other advanced techniques are expected, since the $Q$ matrix remains updated with a correct algorithm. The computational cost of QBIASSR will be discussed in section \ref{sec:experiments}.

Unfortunately, preliminary tests with TOSL+QBIASSR in \mbox{V-REP}, although improving TOSL+SR, have also evidenced a known issue: the chance of falling in cyclic sequences of states before exploring the actions leading to the goal, preventing the robot to efficiently learn the desired task. We believe this is more evident in QBIASSR since it reaches useful sequences of actions sooner than less efficient algorithms; cyclic sequences are also present with softmax occasionally. 

A simple ad-hoc (but task-independent) controller called low-reward-loop evasion (LRLE) has been introduced to solve this problem; it detects whether the agent is selecting low-reward cyclic sequences of actions, and acts by increasing the temperature of the softmax regression, thus favouring the selection of other actions out of the detected low-rewarded sequences. Algorithm \ref{alg:LRLE} describes the implementation of LRLE. 

\begin{algorithm}
	$T$: Boltzmann temperature \\
	$retr$: reduced eligibility traces register (states) \\
	$seq$: FIFO queue of recently visited states. size($seq$) = size($retr$) limited to $[4, states/4]$\\
	$rew$: FIFO queue of recently rewards obtained \\
	Before every softmax selection: \\
	\quad Push the current state $s$ into  $seq$ \\
	\quad \textbf{if} $s$ is redundant in $seq$  \textbf{then} \\
	\quad \quad $a = True$   \Comment{$s$ was visited recently} \\
	\quad Push the current state $R$ into  $rew$ \\
	\quad \textbf{if} avg($rew$) $<0$ \textbf{then} \\
	\quad \quad $b = True$  \Comment{low reward sequence detected} \\
	\quad $u\_seq$ = $seq$ without repeated states \\
	\quad $redundancy$ =  size($seq$) $/$ size($u\_seq$) \\
	\quad \textbf{if} $redundancy > 2$ \textbf{then} \\
	\quad\quad $c = True$    \Comment{repeated behavior detected}
	
	\quad\textbf{if} $a$ \textbf{and} $b$ \textbf{and} $c$ \textbf{then} \\
	\quad \quad $T = T + 0.25\cdot redundancy $ \Comment{evasion rule} \\ 
	\quad \textbf{else}\\
	\quad \quad $T = 1$\\
	\caption{Low-reward-loop evasion (LRLE)}
	\label{alg:LRLE}
\end{algorithm}

Finally, since the probabilities resulting from the softmax regression are affected by the magnitude of the \mbox{$Q$-values} involved (which, in turn, come form rewards) and our goal is to build a RL method independent on the hand-crafted parameters for multiple tasks, we also propose to apply the action selection over a normalized vector. We have employed the theoretical maximum value of $Q$ as a reference for such normalization, being the normalized vector:
\begin{align}
\quad Q(s)_{normalized} = Q(s)\cdot \dfrac{100}{Q_{max}} 
\label{eq:qnorm}
\end{align}
For any temporal difference error seen above, $Q_{max}$ can be obtained from equation \ref{eq:qlearning}  by calculating the limit of $Q$ for infinite steps, resulting in equation \ref{eq:qmax}.

\begin{align}
Q_{max} = \dfrac{R_{max}}{1-\gamma}
\label{eq:qmax}
\end{align}

After that, the learning process will not be affected by the absolute hand-crafted rewards, which contributes to the independence between the decision-making process and the task to learn.

\section{Implementation: the RL-ROBOT software framework}\label{sec:implementation}

For this work we have developed a RL Python-based software framework, which is focused on performing experiments with a variety of robots and environments for different robotic tasks, and that includes TOSL and QBIASSR along with other standard RL algorithms and action selection techniques. The framework, called \mbox{RL-ROBOT}, can be used stand-alone, with the physically realistic robot simulator \mbox{V-REP}, or with the Robot Operating System (ROS) \cite{ROS}, providing then a link to real robots. 

Related works in the RL software arena include several successful tools such as RL-Glue \cite{rl-glue}, RLPy \cite{RLPy}, and the robot-oriented rl-texplore-ros-pkg \cite{hester2013texplore}, widely used for testing RL algorithms and generalization approaches in several environments. The recent advances in deep learning and deep reinforcement learning (DRL) with software libraries such as Tensorflow \cite{tensorflow2015-whitepaper} has also accelerated the development of DRL applications for high-dimensional tasks, mostly based on learning from the input images of a camera, trained by teacher, and sometimes transferred from simulations to real robots. A recent notable contribution, released while this work was in progress, was OpenAI Gym \cite{openAI}, which offers a growing suite of environments for RL, including robotic simulations.

\mbox{RL-ROBOT} provides a framework for robotics researchers that have minimum RL knowledge, and it is ready to perform experiments just by setting their parameters and creating new tasks. A single module serves to specify the parameters of the experiment, including the type of environment, task id, speed rate, repetitions, episodes, steps, algorithm id and its parameters, action selection strategy id, output options (files, charts), etc. Each task module contains the definition of input variables or features (later codified as states automatically), output variables (actions), set of rewards, and the physical devices of the robot (laser rangefinders, motors, etc.). The definition of input variables instead of states, a distinctive feature of our framework, besides of easing the implementation of the task, is used by the QBIASSR algorithm to determine which similar states of $Q$ will influence the decision-making process of a specific state $s$. 

\begin{figure}[h!]
	\centering
	\includegraphics[width=0.48\textwidth]{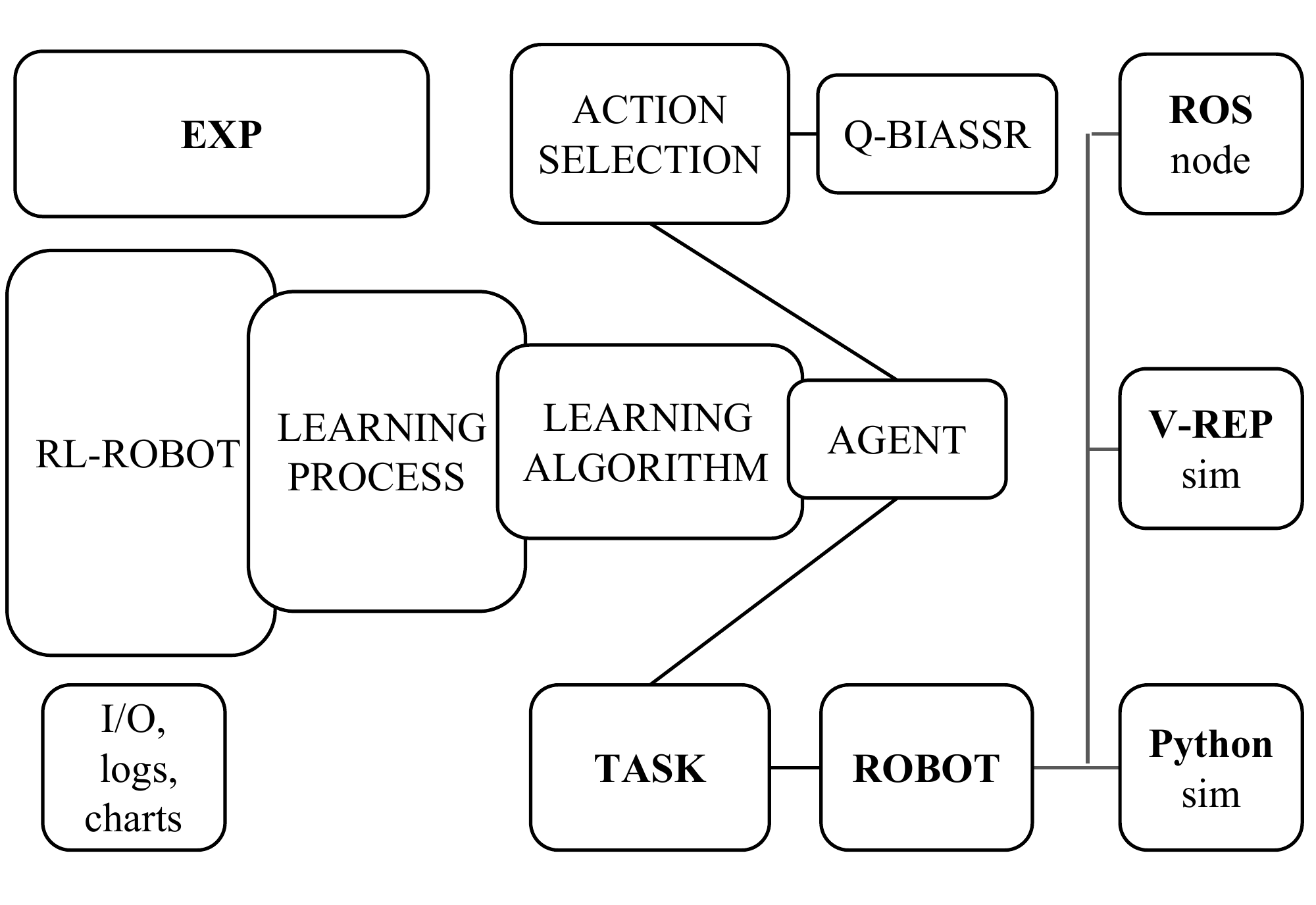}
	\caption{RL-ROBOT software architecture.}
	\label{fig:rl-mapir-diagram}
\end{figure}

Another feature of \mbox{RL-ROBOT} that sets it aside existing RL frameworks is the independence between the abstract learner, portrayed by the artificial agent, and the perceptual robot. We consider that any RL task must define the information of the later; thus a change in the robot, a device, or any input-output variable results in a different task. At the beginning of the learning process, the agent automatically structures itself to connect the learning process with the desired task, no matter the RL algorithm, the action selection technique, the environment, or any hand-crafted definition of the task to learn. The general architecture of \mbox{RL-ROBOT} is summarized in figure \ref{fig:rl-mapir-diagram}.

\mbox{RL-ROBOT} also includes two built-in generic robots: a mobile base with a laser rangefinder, and a 3DOF arm. They can be used separately or together as a single robot (a mobile manipulator). Modules \texttt{V-REP sim} and \texttt{ROS node} implement links with \mbox{V-REP} and ROS, only executed if required, thus connecting our framework with both realistic simulators and real robots. 

\begin{figure}[h!]
	\centering
	\includegraphics[width=0.35\textwidth]{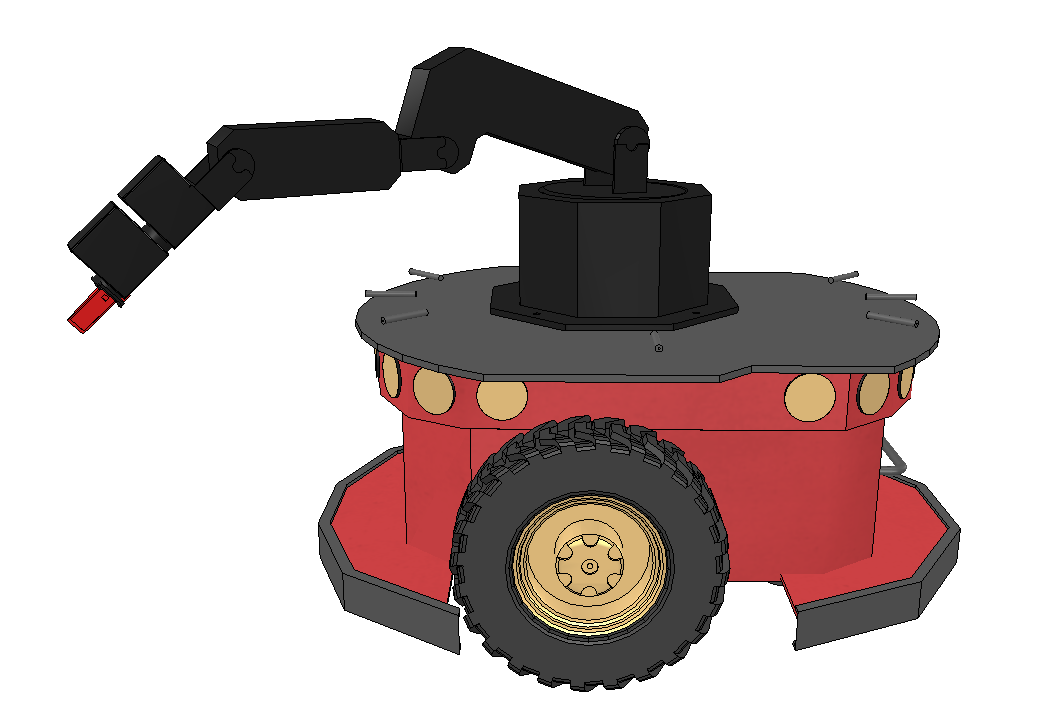}
	\caption{V-REP implemented robot: 3-DX Pioneer mobile base with a WidowX arm.}
	\label{fig:3dx_and_widowx}
\end{figure}

The specific robot that has been used for our \mbox{V-REP} simulations is a Pioneer 3-DX mobile base with 8 laser pointers and a WidowX arm (figure \ref{fig:3dx_and_widowx}). These models have been employed for the simulated experiments of section \ref{sec:experiments}. Since communications with the simulator are via TCP/IP socket using the \mbox{V-REP} remote API, some additions were needed to avoid the influence of network delays and to guarantee the reproducibility of the learning processes in different machines; they include threaded rendering, streaming sensory values, and a limited speed-up ratio of 3 (minimum: Intel Core i3-3110M processor). Physical parameters and scenarios were also tuned to ensure reproducibility. 

A ROS node is automatically launched only when a real robot is required. The node is subscribed to \texttt{odom} and \texttt{laser\_scan} topics, and publishes to \texttt{cmd\_vel} topic. Modules \texttt{V-REP sim} and \texttt{ROS node} can be extended with minimal modifications to connect \mbox{RL-ROBOT} with other robots, as well as other environments implemented in Python.

Three sets of tasks are included in the framework; they have been used for the experiments of section \ref{sec:experiments}: 
\begin{enumerate}
	
	\item[(i)]  wandering (non-episodic): the robot learns to wander avoiding obstacles. The input variables are a reduced set of laser measures around the robot, each distance being discretized within a range. The two output variables are the speeds of the wheels. Positive rewards are received if the robot advances above a distance threshold and negative ones if it collides (where frontal collisions yield a highly negative reward). 
	
	\item[(ii)]  2D mobile navigation (episodic): the robot must reach a static point in the scenario. The input variables here extend the wandering ones  with the discretized pose $(X,Y,\theta)$ of the robot. Rewards maintain the colliding penalties and include positive and negative rewards according to the distance approached to or moved away from the target, respectively.
	
	\item[(iii)]  3D arm motion (episodic): a 3DOF arm must reach an object on a table. The input variables are the $(X,Y,Z)$ positions of the gripper and the object. The outputs are the speeds of up to 3 joints. Positive and negative rewards are received if the gripper approaches to or moves away from the object, respectively.
\end{enumerate}

\begin{figure}[h!]
	\centering
	\includegraphics[width=0.48\textwidth]{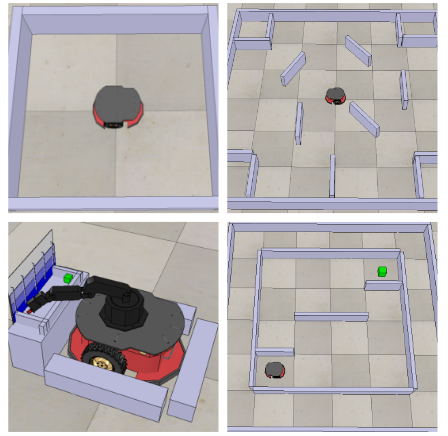}
	\caption{V-REP scenarios for our experiments. top) 2x2m and 6x6m with obstacles, for wandering. bottom-left) 32x12cm for 3D arm motion. bottom-right) 4x4m with obstacles for 2D mobile navigation.}
	\label{fig:scenarios}
\end{figure}

The particular tasks implemented are listed in table \ref{tab:tasks}, while figure \ref{fig:scenarios} shows the scenarios designed in \mbox{V-REP} for the above tasks. \mbox{RL-ROBOT} has been released as an open source project, Python PEP8 style, available on Github \cite{angel_martinez_tenor_2016_166555}. 

\begin{table*}[h!]
	\small\sf\centering
	\caption{Tasks implemented for the learning experiments.}
	\begin{tabular}{p{0.22\textwidth}p{0.2\textwidth}p{0.15\textwidth}p{0.12\textwidth}}
		\toprule
		Set & Name & States & Actions \\
		\midrule
		Wandering & wander-simple & 4 & 4 \\
		&wander-1K & 1024 & 9  \\
		&wander-4K & 4096 & 25 \\
		\midrule
		3D arm motion &3D-arm-1K & 1024 & 9 (2 joints) \\
		&3D-arm-4K & 4096 & 27 (3 joints)\\
		\midrule
		2D mobile navigation &2D-navigation-1K & 1024 & 9  \\
		\bottomrule
	\end{tabular}
	\label{tab:tasks}
\end{table*}

\section{Experiments and discussion}\label{sec:experiments}

The experiments of this paper have been conducted in \mbox{RL-ROBOT} with both simulated and real robots. Periods of time of up to 60 minutes were used in simulated experiments so as to be reproducible on real robots. A learning curve showing the evolution of the average reward obtained over time (steps) has been found suitable as both a stable measurement of the performance of the learning process and a reliable indicator for comparing different techniques. For episodic tasks, the evolution of the average reward of the last episode was used as indicator instead of the classical evolution per episode, since the number of episodes required were low. The results include the average learning curve of several repetitions, along with an analysis of variance and a post-hoc Tukey test \cite{tukey1949comparing} for each task in order to assess the significance of the conclusions.

\subsection{Sample-modeled and V-REP simulated experiments}

Simulated experiments have been performed for both sample models and realistic simulators. Sample-modeled experiments are based on Markovian models consisting of $T(s,a,s')$ and $R(s,a,s')$ that are built offline after exploring the tasks in the \mbox{V-REP} simulator. On the contrary, in the realistic simulated experiments, each learning step has been performed in one second of simulation time directly in \mbox{V-REP}. The total time for each learning process ranges from 1 to 60 minutes (3600 steps) according to the task to learn. 

\begin{figure}[h!]
	\centering
	\includegraphics[width=0.48\textwidth]{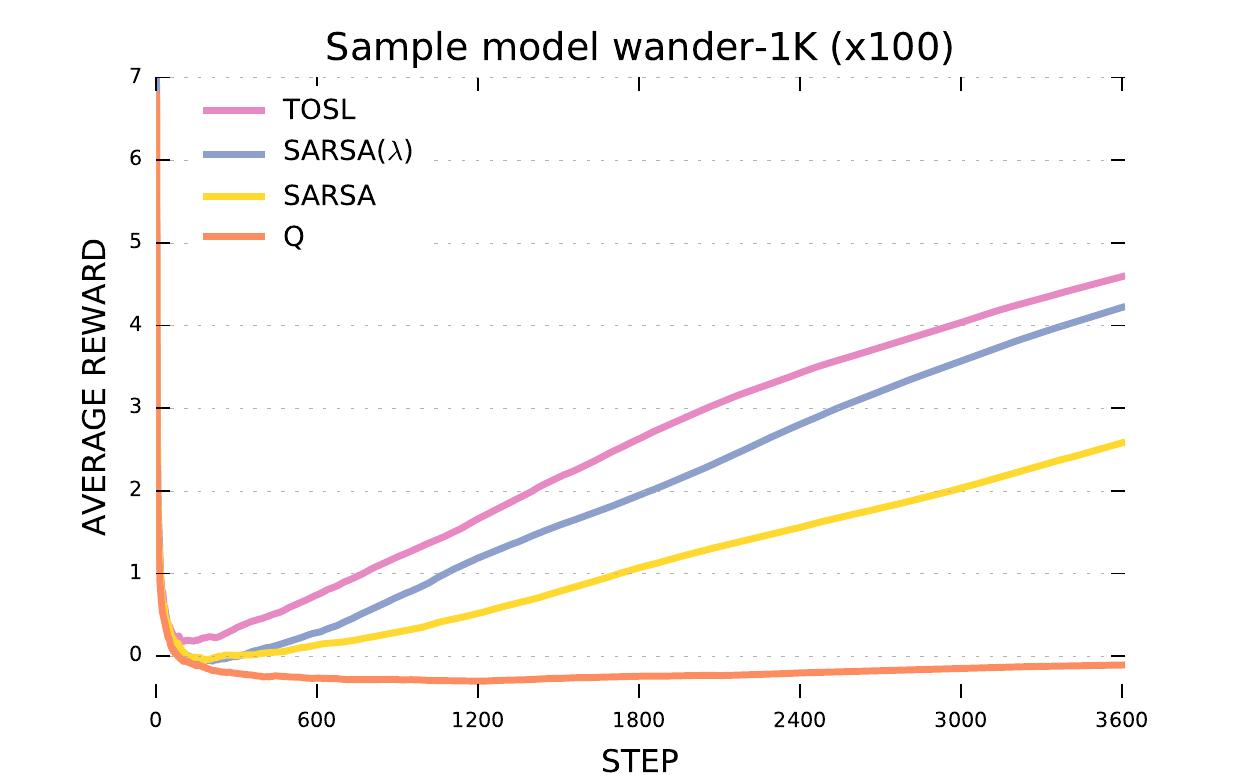}
	\includegraphics[width=0.48\textwidth]{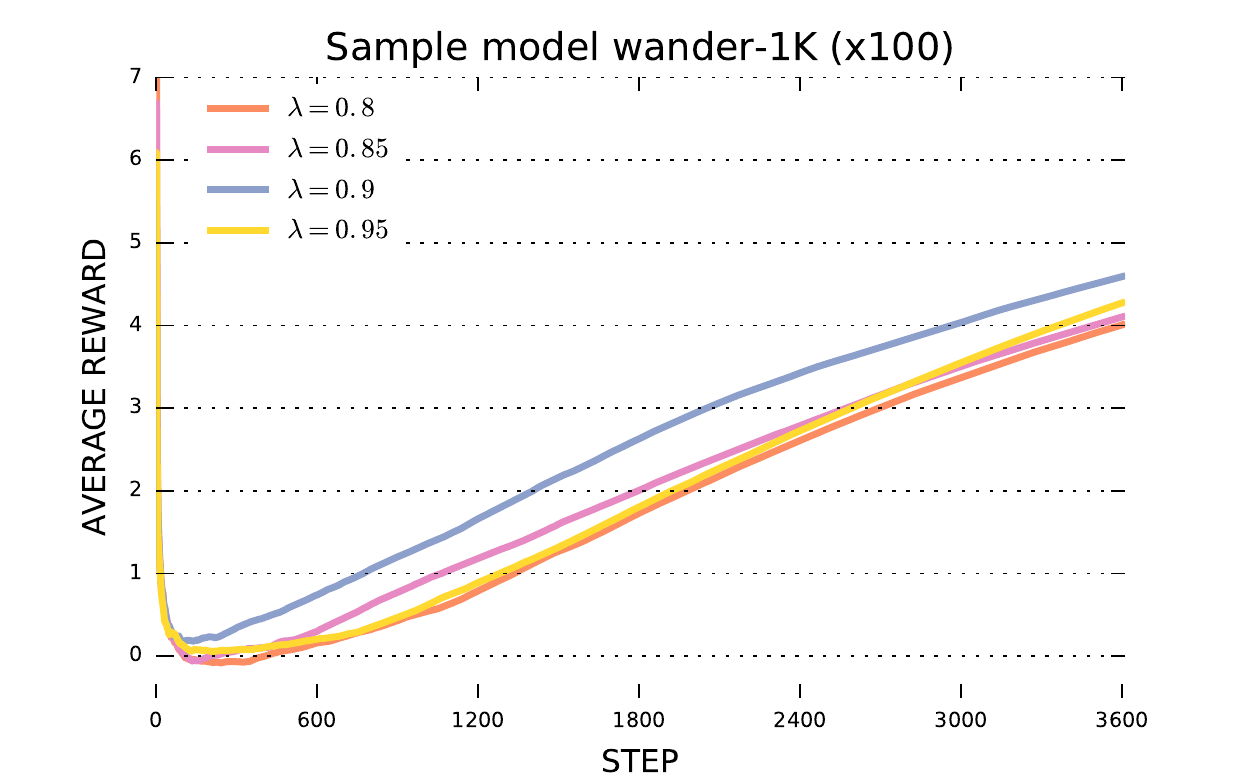}
	\caption{Sample-modeled preliminary tests for the tasks wander-1K in a 6x6m scenario with obstacles. top) true online SARSA($\lambda$) vs SARSA($\lambda$) vs SARSA vs Q-learning. bottom) true online SARSA($\lambda$) for diverse ET factor values.}
	\label{fig:core-algorithms}
\end{figure}

The preliminary tests described in section \ref{sec:approach} have been performed to obtain a basic algorithm valid for multiple tasks, chosen from existing RL ones; this led to TOSL, as it has been explained. Both sample-modeled and \mbox{V-REP} simulations have been used with multiple tasks in that stage. Figure \ref{fig:core-algorithms} shows some results that exemplify the selection of this algorithm and its parameter $\lambda$.

Once TOSL is taken as the basic core algorithm, the rest of our simulated experiments have been focused on the performance of true online SARSA($\lambda$) with Q-biased softmax regression (TOSL+QBIASSR) in comparison to true online SARSA($\lambda$) with softmax regression (TOSL+SR) and \mbox{Q-learning} with softmax regression (Q+SR). Two sets of experiments have been carried out here: offline \mbox{sample-modeled} simulations and \mbox{V-REP} simulations. Three algorithms have been evaluated for the tasks described in section \ref{sec:implementation}, with different number of states and actions, inside the scenarios that were shown in figure \ref{fig:scenarios}. 

\begin{figure}[h!]
	\centering
	\includegraphics[width=0.48\textwidth]{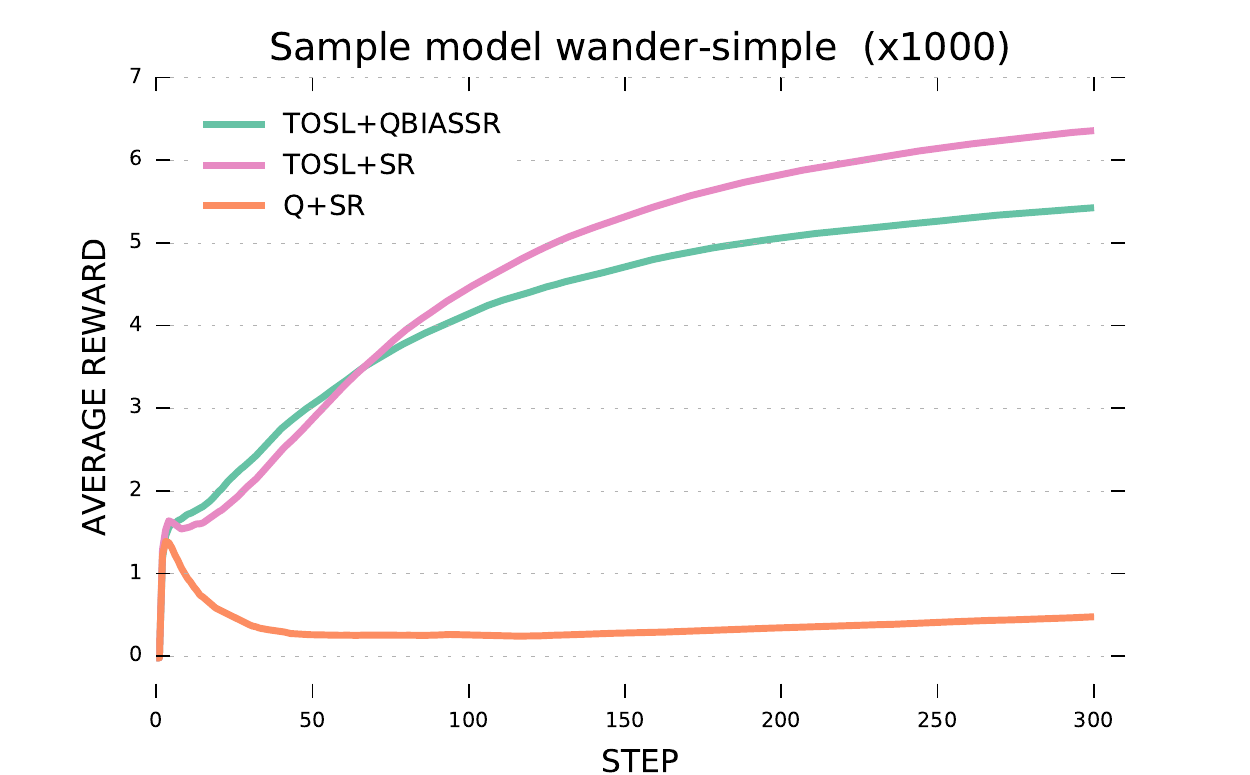}
	\includegraphics[width=0.48\textwidth]{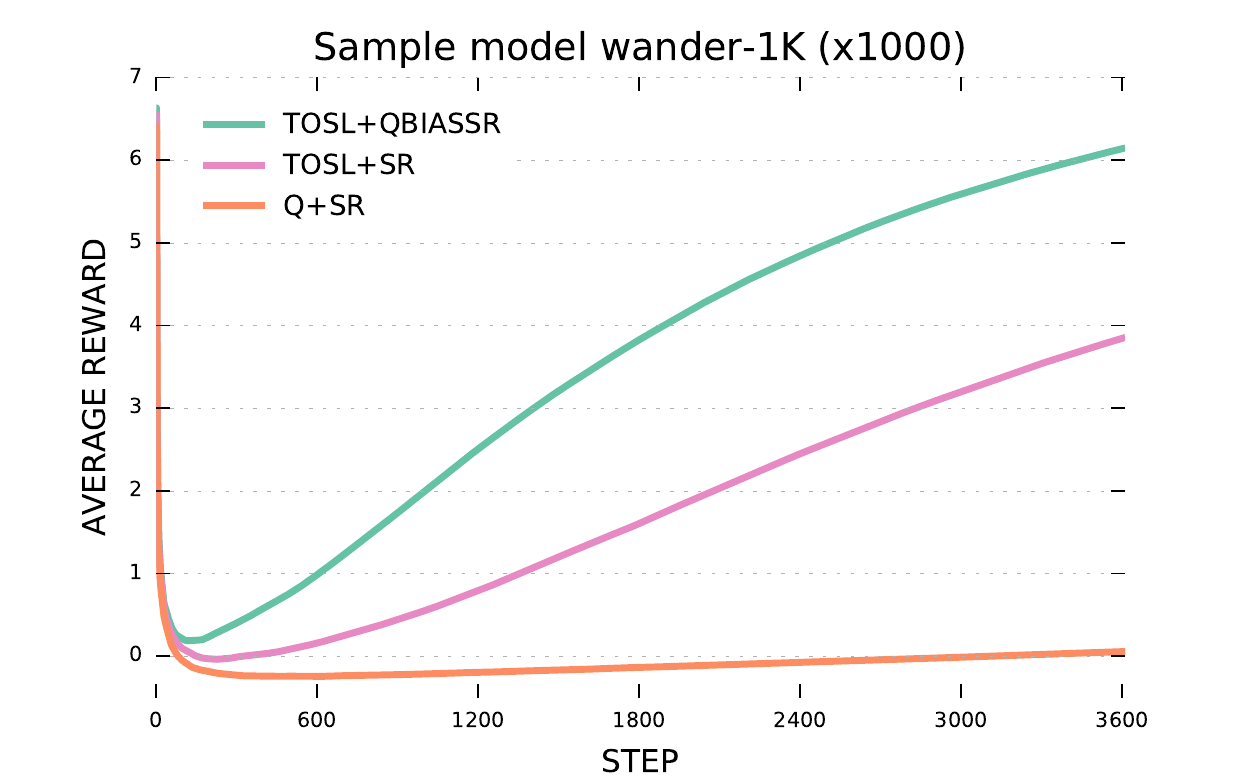}

	\includegraphics[width=0.48\textwidth]{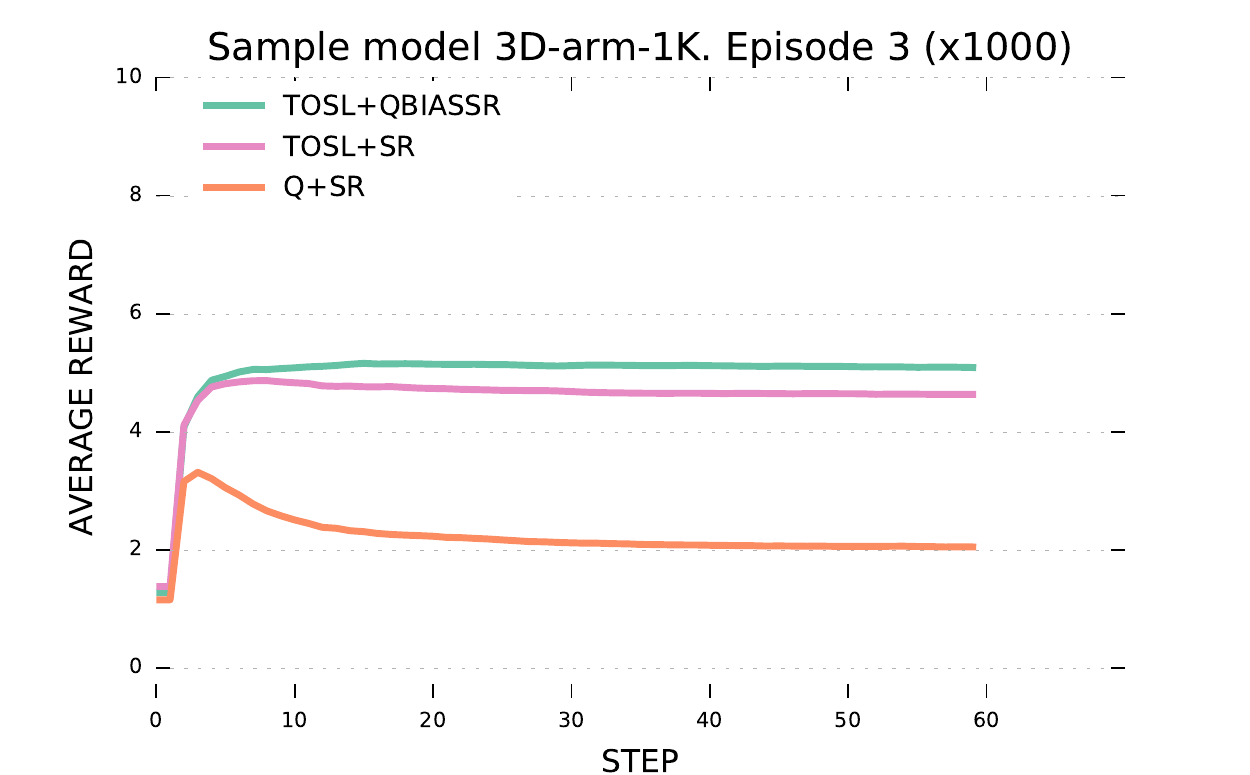}
	\includegraphics[width=0.48\textwidth]{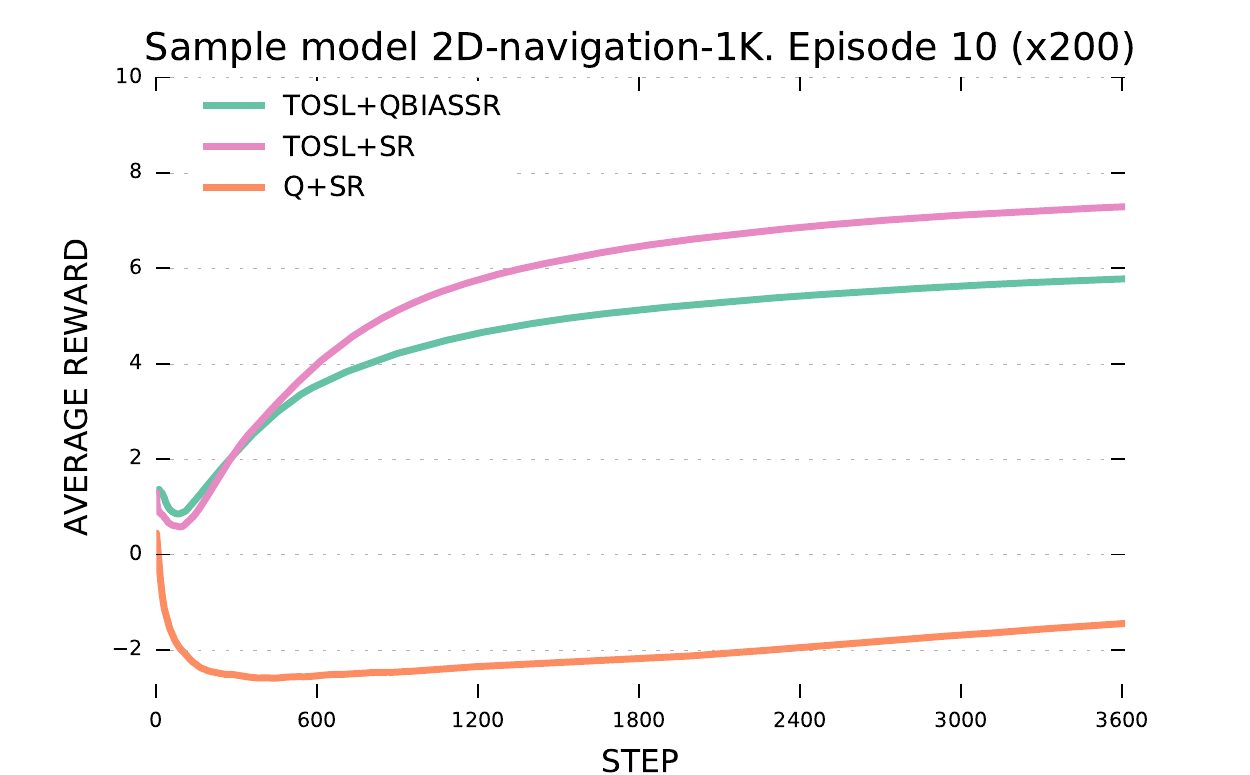}
	\caption{Sample-modeled simulations. TOSL+QBIASSR vs TOSL+SR vs Q+SR learning four robotic tasks.}
	\label{fig:model-charts}
\end{figure}

The results of the learning processes in Markovian sample models have been averaged from 1000 repetitions for wandering and 3D arm motion, and from 200 repetitions for 2D mobile navigation. Figure \ref{fig:model-charts} shows the learning curves for a set of sample-modeled tests. Experiments involving more than 2K states where not performed due to memory limitations when creating the reward function from datasets with variable rewards. The results show that TOSL+QBIASSR only outperformed TOSL+SR in two of the four tasks. Different sample models extracted from V-REP were tested with unexpected diverse results. This led us to focus on realistic V-REP simulations instead of the sample-modeled ones.

\begin{figure}[h!]
	\centering
	\includegraphics[width=0.48\textwidth]{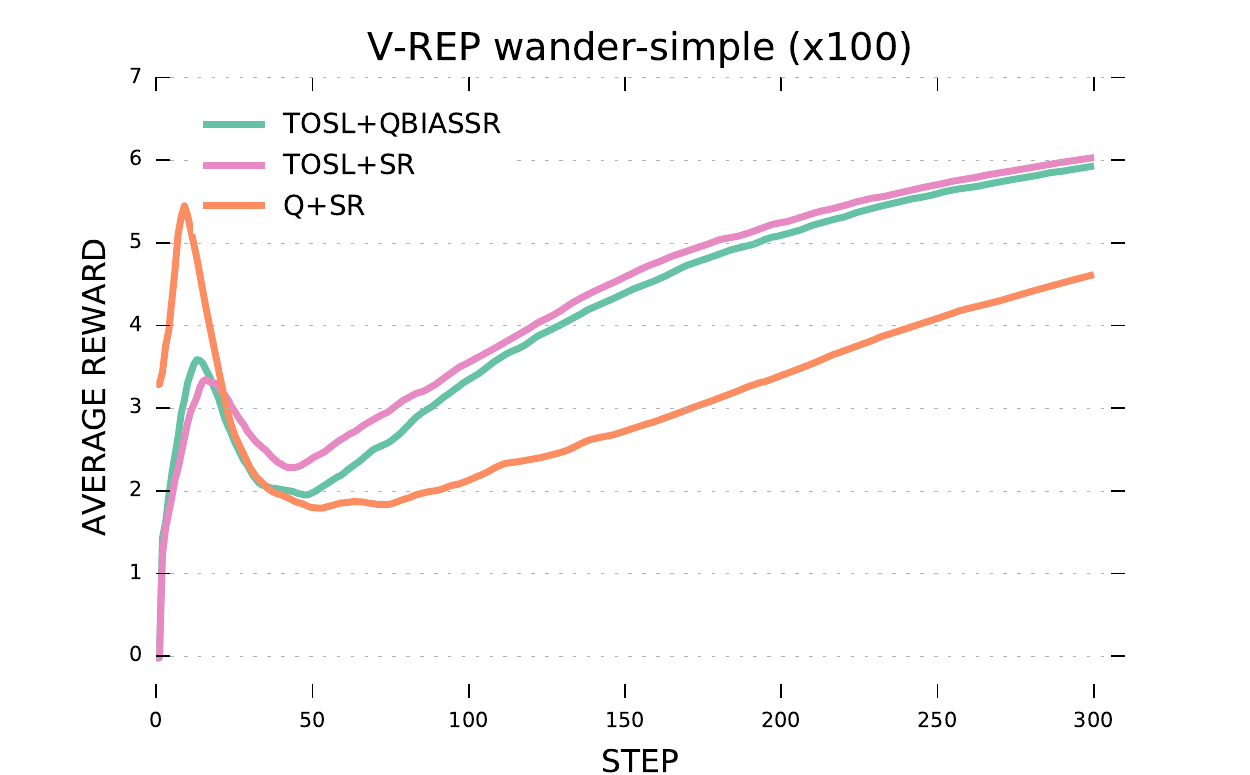}
	\includegraphics[width=0.48\textwidth]{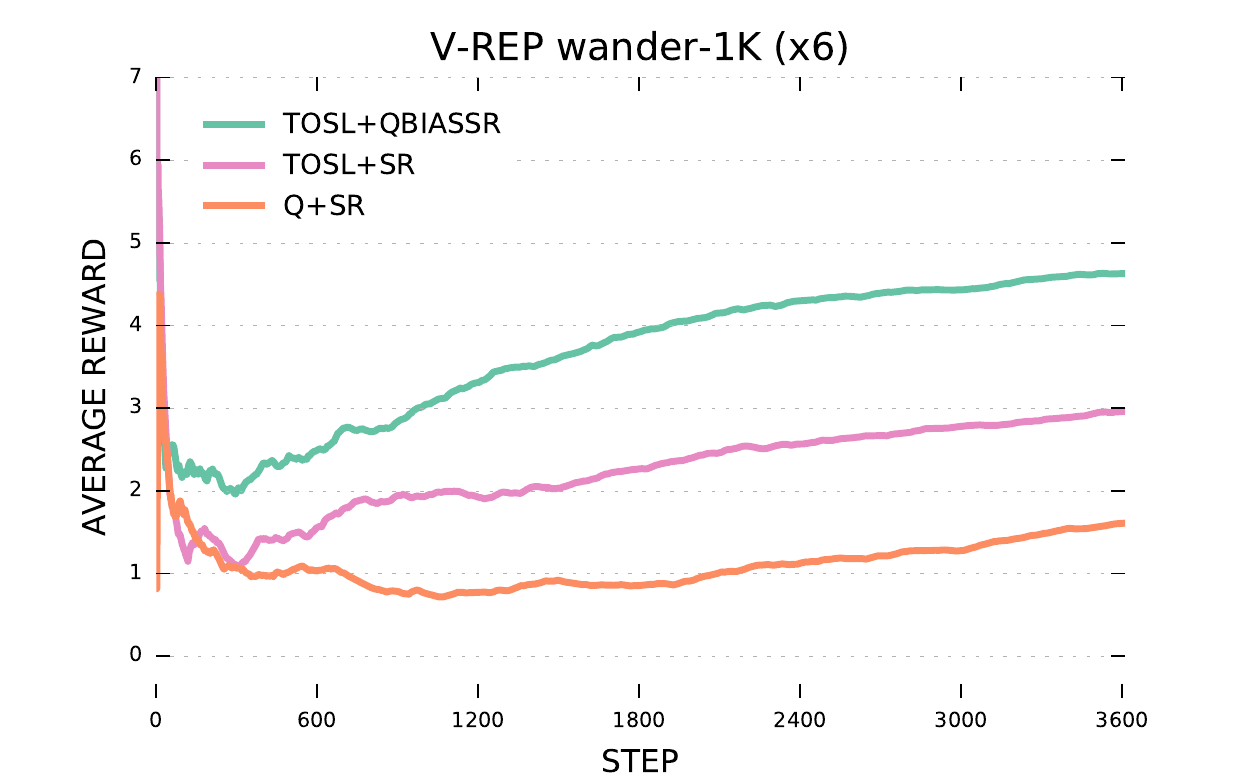}
	\includegraphics[width=0.48\textwidth]{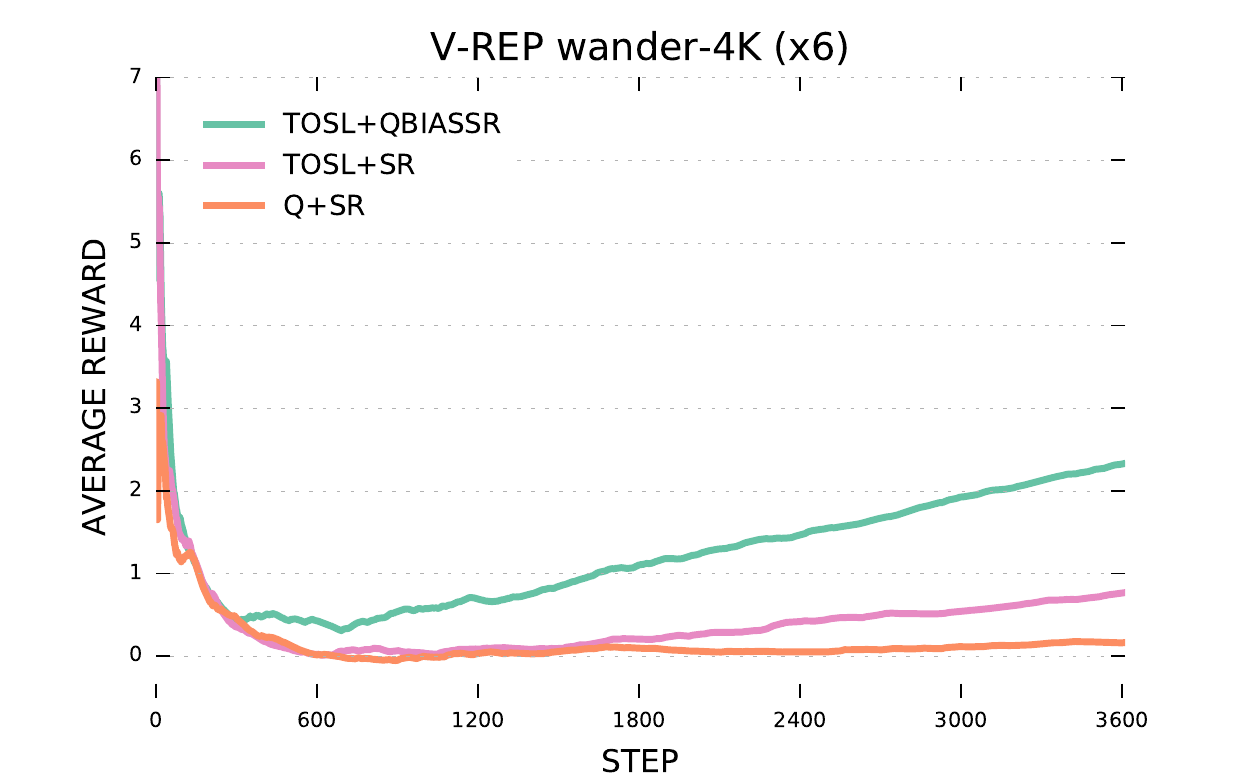}
	\caption{V-REP simulations. TOSL+QBIASSR vs TOSL+SR vs Q+SR learning three wandering tasks.}
	\label{fig:vrep-charts}
\end{figure}

\begin{figure}[h!]
	\centering
	\includegraphics[width=0.48\textwidth]{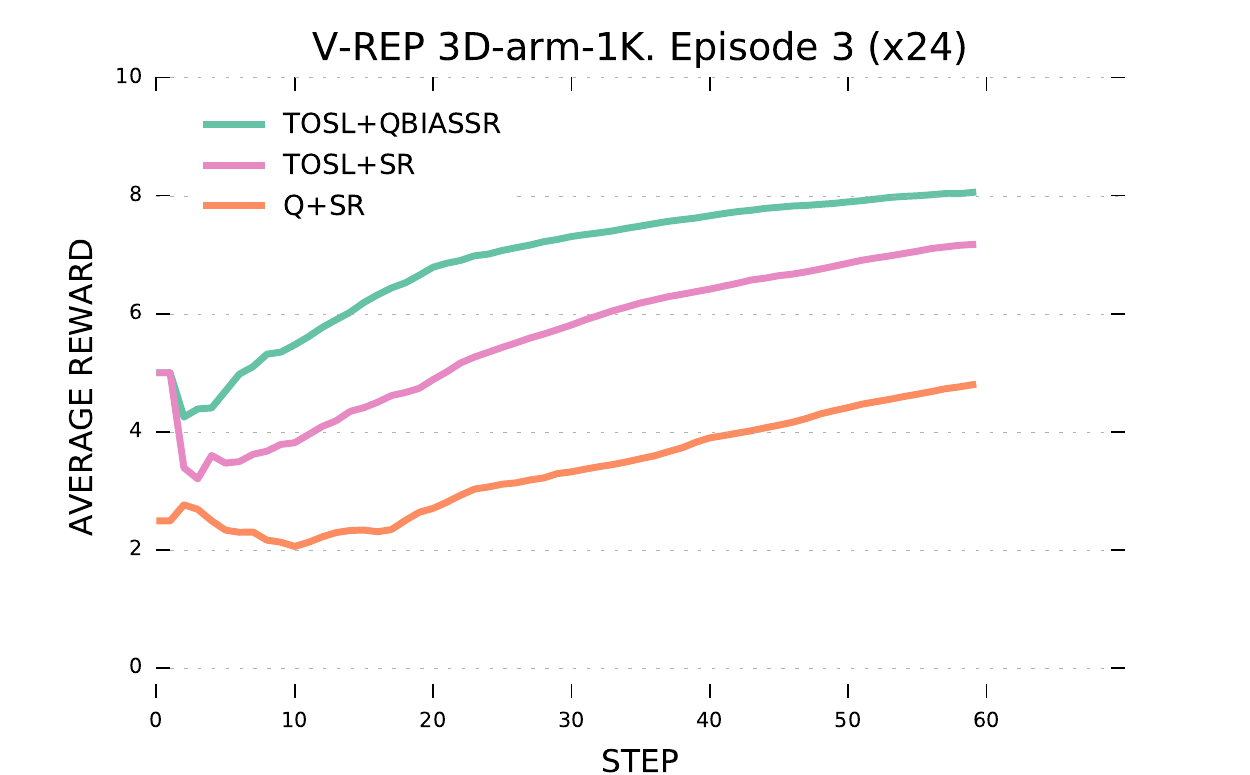}
	
	\includegraphics[width=0.48\textwidth]{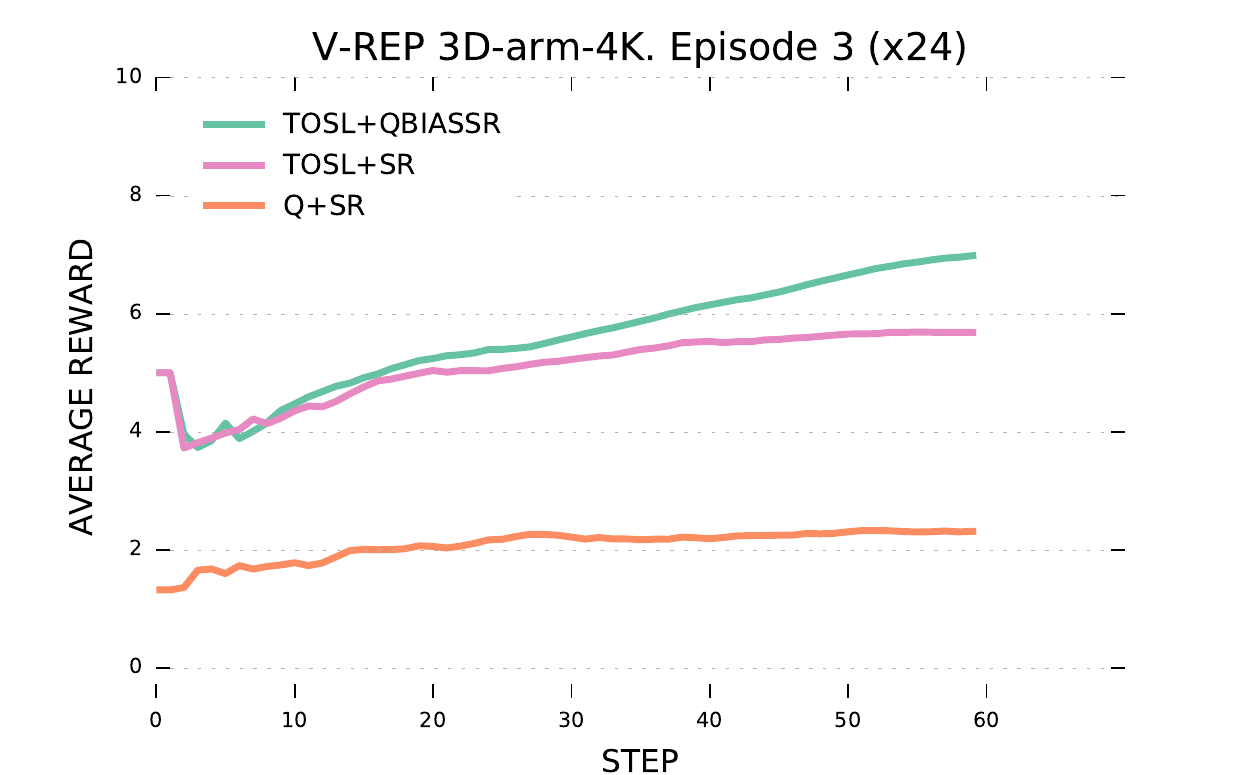}
	\includegraphics[width=0.48\textwidth]{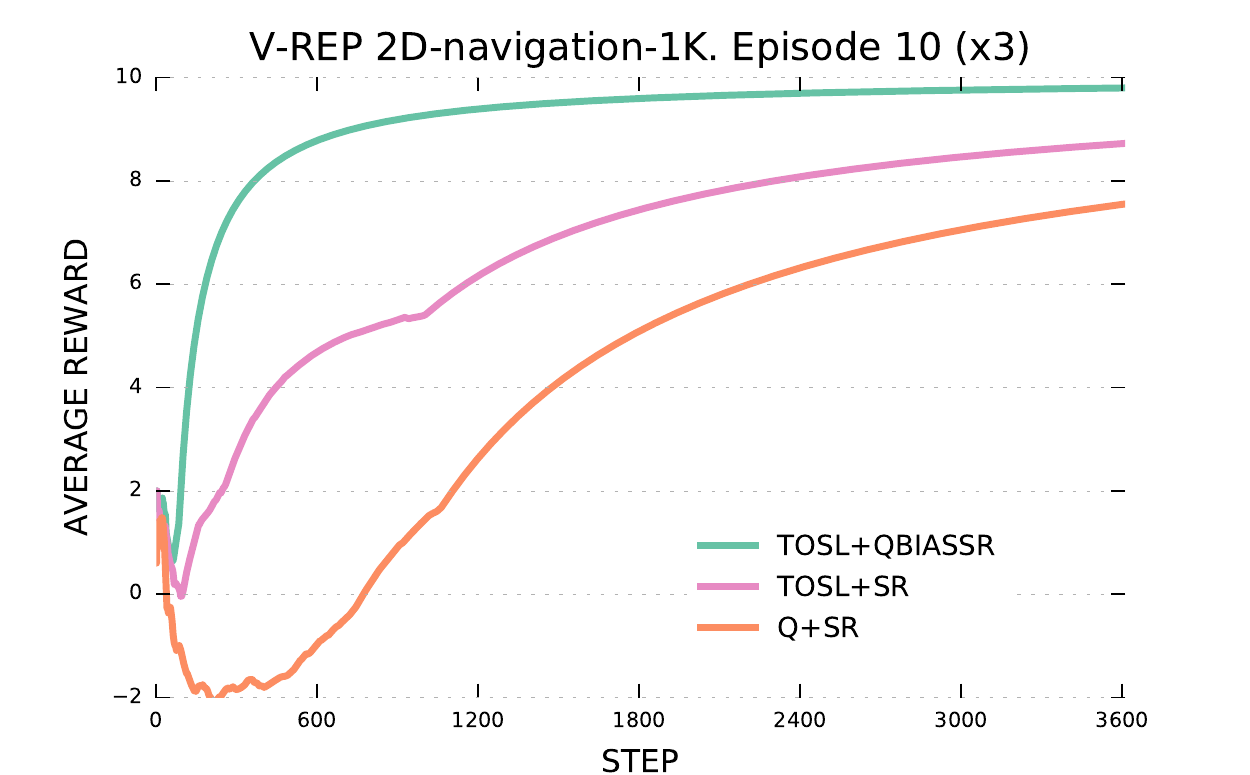}
	\caption{V-REP simulations. TOSL+QBIASSR vs TOSL+SR vs Q+SR learning 3D arm motion and 2D navigation tasks.}
	\label{fig:vrep-charts2}
\end{figure}

The learning experiments with the non-Markovian realistic \mbox{V-REP} simulations gave accurate results. Besides, without the above memory restriction, we could extend the study to the tasks wander-4K and 3D-arm-4K. The resulting learning curves are shown in figures \ref{fig:vrep-charts} and  \ref{fig:vrep-charts2}, where it is shown how the novel TOSL+QBIASSR outperforms TOSL+SR in five of the six tasks tested, never affecting negatively any learning process, nor even the simplest wandering task, where better results than TOSL+SR were unlikely. 

However, an analysis of variance of the average reward obtained in each repetition (table \ref{tab:ANOVA-VREP}) and a post-hoc Tukey test (table \ref{tab:Tukey-VREP})  show that the effect on the learning process of QBIASSR over SR is significant (p$<$0.01) only for two of the six tasks: wander-1K and wander-4K. This is due to the large variance existing in non-wandering tasks (the exploration before getting the high-rewarded goal produces a lot of uncertainty in episodic tasks). Other tasks with smaller variance (i.e., wandering) lead to a clear improvement of QBIASSR in the learning process, shown by the differences in the average curves appreciable in figure \ref{fig:vrep-charts}. We also can confirm that most sample-modeled learning simulations performed worse than the realistic ones, especially the basic \mbox{Q-learning} algorithm, showing poor learning even in the most simple task. This behavior suggests that the Markovian models used in the first place were too simple to accurately represent real robotic environments, and also that non-Markovian real-world and physically realistic simulated environments offer more suitable scenarios than Markovian models for robotic tasks.

\subsection{Real robot experiments}

\begin{figure}[h!]
	\centering
	\includegraphics[width=0.45\textwidth]{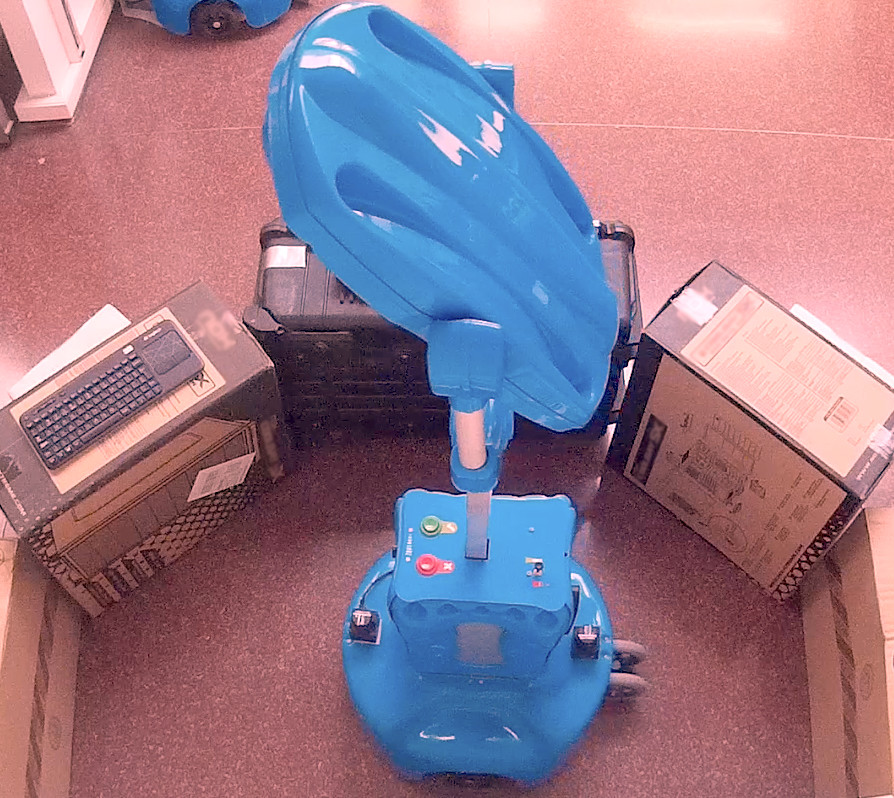}
	\caption{Giraff robot in the wandering task scenario.}
	\label{fig:giraff-scenario}
\end{figure}

The next set of experiments have been conducted with the Giraff robot shown in figure \ref{fig:giraff-scenario}, that is equipped with two differential drive and two caster wheels, and a Hokuyo laser rangefinder. The same tasks designed for the simulated mobile robot can be learned by Giraff just by specifying the physical parameters of the task. \mbox{RL-ROBOT} launches the ROS node in this case, giving access to the real sensors and actuators. 

\begin{figure}[h!]
	\centering
	\includegraphics[width=0.48\textwidth]{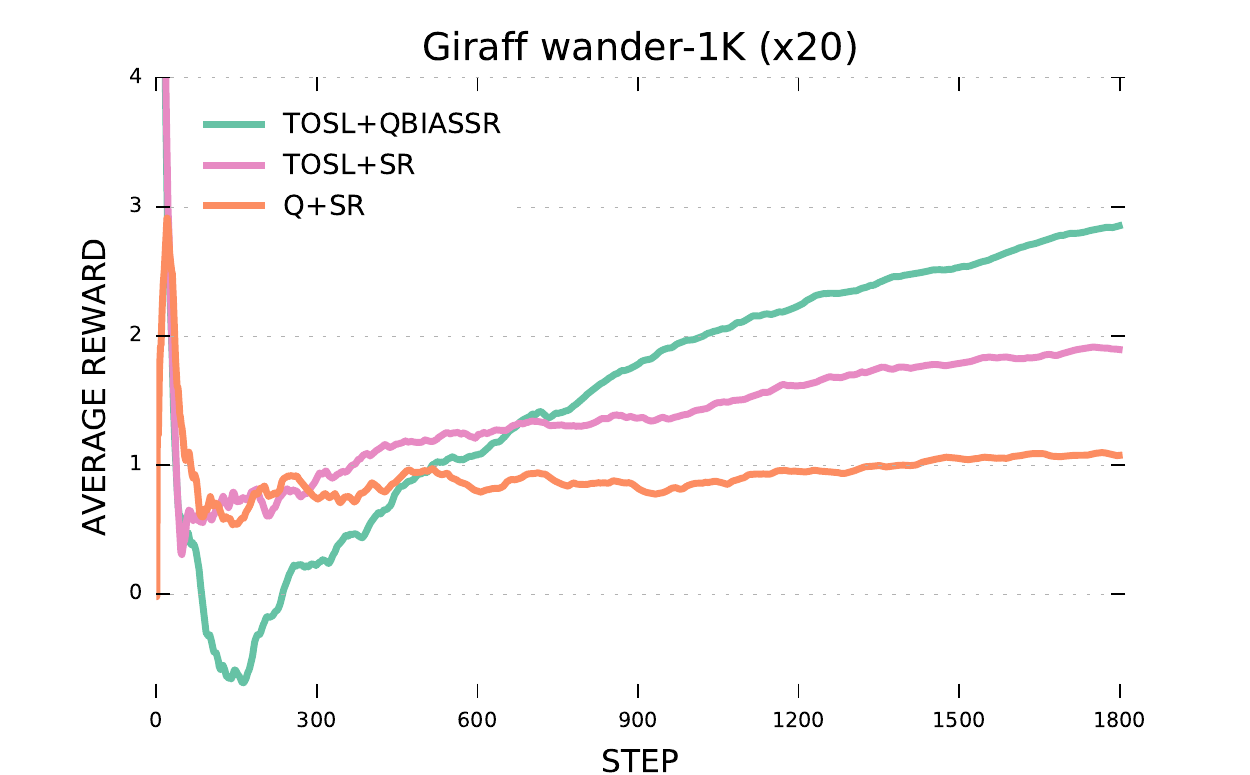}
	\caption{Giraff robot. TOSL+QBIASSR vs TOSL+SR vs Q+SR learning wander-1K task.}
	\label{fig:giraff-results}
\end{figure}

Giraff must learn the task wander-1K in a 1.5x1m hexagonal scenario. Up to 20 learning processes of 30 minutes each have been executed for TOSL+QBIASSR, TOSL+SR, and Q+SR. The results shown in figure \ref{fig:giraff-results} highlight the differences between the evaluated methods.

The learning curve of TOSL+QBIASSR presents a continuous drop (collisions) during the fist 3 minutes (180 steps), after which we consider the robot has learned to perform the task. Moreover, the slope of the average reward curve at the last steps suggests that in 3 minutes, TOSL+QBIASSR learns to wander more effectively than the rest of the algorithms do after 30 minutes. Q+SR, for instance, results in a much flatter learning process. The analysis of variance (table \ref{tab:ANOVA-giraff}) indicates that the effect of the RL method executed by  Giraff is significant, $F(2,57)=16.36$, $p = 2\cdot 10^{-6}$. A Tukey post-hoc test (table \ref{tab:Tukey-giraff}) reveals that TOSL+QBIASSR significantly overcomes TOSL+SR at p$<$0.01, resulting unexpectedly better than the improvement of TOSL over \mbox{Q-learning}.

\subsection{Reduced eligibility traces}
\begin{figure}[h!]
	\centering
	\includegraphics[width=0.48\textwidth]{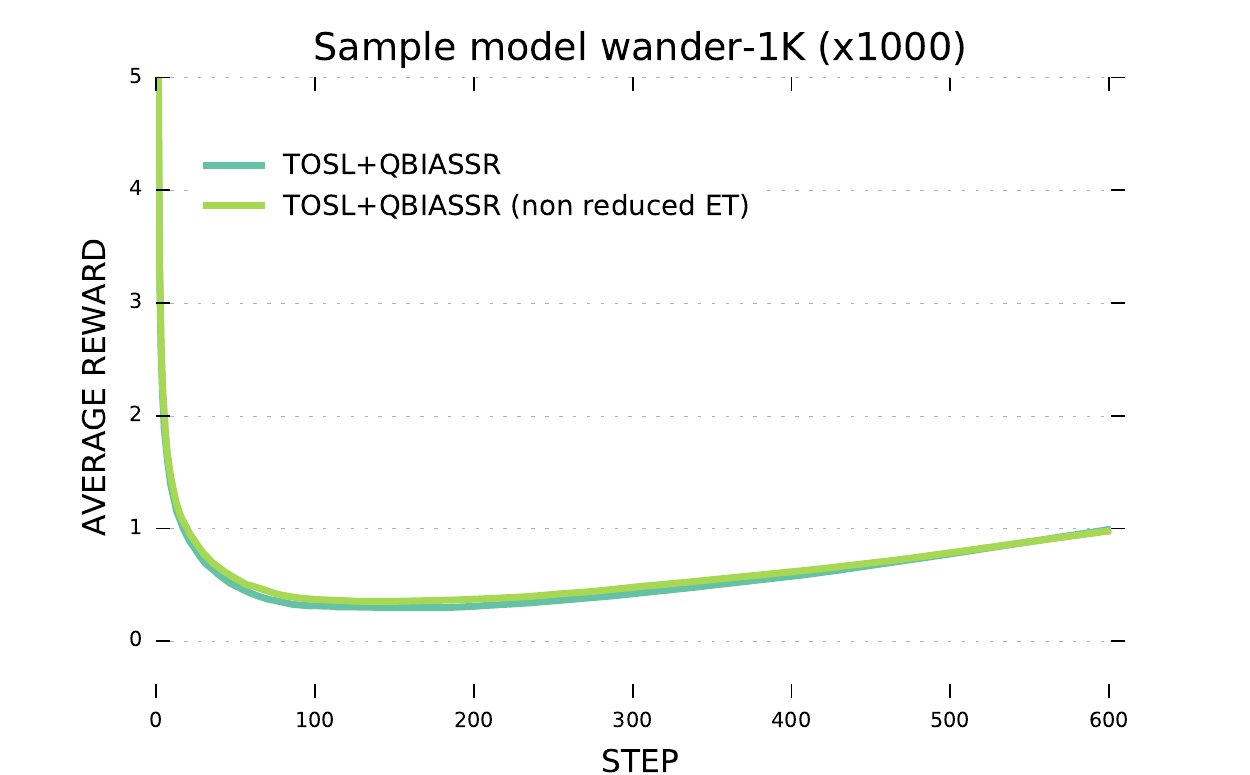}
	\caption{Reduced vs standard ET in sample-modeled wander-1K (10 minutes).}
	\label{fig:reduced-ET}
\end{figure}

A comparison test between the reduced version of eligibility traces (ET) proposed in section \ref{sec:approach} and standard ET has also been performed in sample-modeled simulation to verify that the reduced ET used in this work has a negligible effect over the learning process. This conclusion is clearly shown in figure \ref{fig:reduced-ET}.

\subsection{Computational cost}
Finally, a last set of tests has been carried out to measure the computational cost of the algorithms used in this work. For this purpose the tests were executed in sample-modeled simulation for the wander-1K task. Different RL implementations were tested, executing 30 learning process of 3600 steps each. The measurement of choice to assess the computational impact of the learning algorithms is the average CPU time consumed in each learning step. The results, obtained with RL-ROBOT in an Intel Core i5-4460 processor under Ubuntu 64 bits, are shown in table \ref{comp_cost}. The execution times obtained for TOSL, using the reduced ET discussed in section \ref{sec:approach}, are up to 17 times faster than the standard ET implementation, without affecting the performance of the learning algorithm; they even have the same order of magnitude as the reference times of \mbox{Q-learning} and SARSA, which we consider negligible for RL applications. The computational cost of TOSL (reduced ET) with QBIASSR action selection also results in the same order of magnitude as the above methods. Though QBIASSR uses extensive information from many states, its reduced computational impact has been achieved due to the fact that all the structure of the sets of states biasing a particular state can be generated just by knowing the input variables involved. This allows an efficient implementation, computing most of the QBIASSR algorithm  beforehand, and obtaining a delay of $\approx 1$ second before beginning the learning process, measured in the same conditions as before. For real-time robotic applications, even non-reduced ET could be used with minimal impact on the learning process, since the waiting time for reaching a new state is usually much longer than the one needed for updating $Q$ for all the states.

\begin{table*}[h!]
	\small\sf\centering
	\caption{Computational cost. Sample-modeled wander-1K}
	\begin{tabular}{p{0.2\textwidth}p{0.2\textwidth}p{0.25\textwidth}p{0.25\textwidth}}
		\toprule
		RL algorithm & Reduced ET & Avg CPU time per step (s) & Total CPU time (3600 steps) (s)   \\
		\midrule
		Q+SR & - & $0.19\cdot 10^{-3}$ & $0.70$ \\
		SARSA+SR & - & $0.20\cdot 10^{-3}$ & $0.72$\\
		SARSA($\lambda$)+SR & Yes & $0.31\cdot 10^{-3}$ & $1.13$\\
		TOSL+SR & Yes & $0.31\cdot 10^{-3}$ & $1.13$\\
		TOSL+QBIASSR & Yes & $0.45\cdot 10^{-3}$ & $1.62$\\
		TOSL+SR & No & $5.32\cdot 10^{-3}$ & $19.17$\\
		\bottomrule
	\end{tabular}
	\label{comp_cost}
\end{table*}

\section{Conclusions and future work}\label{sec:conclusions} 

We have demonstrated in this paper that state-of-the-art value-iteration-based RL algorithms that evolved from classical \mbox{Q-learning}, such as TOSL, can be effective when applied in multiple real tasks, something still unachieved in robotics. For that, we have found suitable common, basic parameters for TOSL, and we have introduced a novel exploration technique, QBIASSR, which improves the classical softmax action selection. QBIASSR takes advantage of the physical input variables used by the robot, thus the action selection can be influenced by other previously visited states with similar inputs. A complementary low-reward-loop evasion algorithm has been added to prevent local optima sequences. The combined algorithm TOSL+QBIASSR has the advantage of being both task-independent and compatible with most advanced RL techniques for improving learning in higher dimensional tasks, with a negligible computational cost.

A comparative study with other RL techniques has been performed, including both non-episodic and episodic robotic tasks. Realistic simulations and real robot experiments reveal that the learning processes with the novel TOSL+QBIASSR outperform those with TOSL+SR in most tasks, being equivalent to TOSL+SR in the worst case scenario. An additional result has been that Markovian sample-modeled learning processes are unrealistic for robotic tasks because of the gap of the performance detected between V-REP simulations and the models extracted from them.

The \mbox{RL-ROBOT} software framework has also been introduced in this paper. It is an open source project designed to ease the implementation of new tasks for both simulated and real robots, including \mbox{V-REP} and ROS interfaces. Just by defining the physical variables and parameters involved, each task provides structured perceptual information of the environment to the abstract agent regardless the learning method used. \mbox{RL-ROBOT} contributes a novel implementation that achieves independence between the abstract and perceptual aspects involved in RL in robotics.

Future work includes the addition of advanced techniques for boosting the learning process in robotics, e.g., by constructing a model (\mbox{model-based} RL), such as prioritized sweeping, by human interaction as teaching, and by transferring \cite{taylor2009transfer}, \cite{barrett2010transfer} and \cite{bianchi2015transferring} from simulations to real robots. We also believe that these techniques will contribute to decrease the variance of the learning processes seen here, which sometimes have produced inconclusive results, especially in episodic tasks.

TOSL+QBIASSR can be extended for more practical high-dimensional tasks as well, instead of the low-dimensional tasks shown here. Advanced generalization methods for RL include a wide variety of function approximators, techniques for continuous states and action spaces, Bayesian methods, etc. A suitable generalization exploiting the basis of QBIASSR could be hierarchical RL \cite{barto2003recent} and \cite{botvinick2012hierarchical}, where complex tasks could be formed by adding more input variables to existing tasks. However, for maintaining a strong independence between the task and the RL method, we contemplate the recent advances of DRL as promising techniques to extend this work to more complex robotic tasks.

Finally, we plan to integrate the \mbox{RL-ROBOT} framework with the OpenAI gym toolkit, where TOSL+QBIASSR and the upcoming changes can be evaluated and compared for a wide variety of tasks.

\begin{table*}[h!]
	\small\sf\centering
	\caption{ANOVA: Learning techniques on V-REP sim}
	\begin{tabular}{p{.2\textwidth}p{.16\textwidth}p{.12\textwidth}p{.04\textwidth}p{.11\textwidth}p{.055\textwidth}p{.1\textwidth}}
		\toprule
		task            &source             &sum of squares   &df    &mean square   &F         & p-value \\
		\midrule
		&between groups     & 125.41            & 2        & 62.7        & 39.09    & $8.88\cdot10^{-16}$\\
		wander-simple      &within groups      & 476.37            & 297      & 1.60     \\  
		&total              & 601.78            & 299                  \\
		\midrule
		&between groups      & 27.47             & 2         & 13.74        & 31.29    & $4.44\cdot10^{-6}$\\
		wander-1K  &within groups       & 6.58              & 15       & 0.44     \\  
		&total               & 34.06             & 17                  \\
		\midrule
		&between groups      & 14.96              & 2          & 7.48        &  32.05   & $3.84\cdot10^{-6}$\\
		wander-4K  &within groups       & 3.50               & 15           & 0.23      \\  
		&total               & 18.46              & 17                  \\
		\midrule
		&between groups      & 136.21       & 2          & 68.10     &  10.30   & 0.0001\\
		3D-arm-1K    &within groups       & 456.23       & 69         & 6.61      \\  
		&total               & 592.43       & 71 \\
		\midrule
		&between groups      & 278.65       & 2          & 139.33     &  23.17   & $2.01\cdot10^{-8}$\\
		3D-arm-4K   &within groups       & 414.97       & 69         & 6.01      \\  
		&total               & 693.62       & 71                    \\
		\midrule
		&between groups      & 7.62       & 2           & 3.81     &  4.15   & 0.070\\
		2D-navigation-1K          &within groups       & 5.51       & 6           & 0.92      \\  
		&total               & 13.13      & 8                    \\
		\bottomrule
	\end{tabular}
	\label{tab:ANOVA-VREP}
\end{table*}

\begin{table*}[h!]
	\small\sf\centering
	\caption{Post-hoc Tukey HSD test on V-REP sim}
	\begin{tabular}{p{.2\textwidth}p{.26\textwidth}p{.14\textwidth}p{.11\textwidth}p{.12\textwidth}}
		\toprule
		task            &pair                           & HSD Q statistic   & HSD p-value    & HSD inference \\
		\midrule
		& TOSL+QBIASSR  vs TOSL+SR      & 0.80	        & 0.820     \\
		wander-simple      & TOSL+QBIASSR  vs Q+SR         & 10.41	        & 0.001	    & p$<$0.01 \\
		& TOSL+SR vs Q+SR               & 11.21	        & 0.001	    & p$<$0.01  \\          
		\midrule
		& TOSL+QBIASSR  vs TOSL+SR      & 6.17	        & 0.002     & p$<$0.01 \\
		wander-1K    & TOSL+QBIASSR  vs Q+SR       & 11.17	        & 0.001	    & p$<$0.01 \\
		& TOSL+SR vs Q+SR               & 5.00	        & 0.008	    & p$<$0.01  \\ 
		\midrule
		& TOSL+QBIASSR  vs TOSL+SR     & 7.92	        & 0.001     & p$<$0.01 \\
		wander-4K    & TOSL+QBIASSR  vs Q+SR         & 10.97	        & 0.001	    & p$<$0.01 \\
		& TOSL+SR vs Q+SR               & 3.05	        & 0.111	      \\ 
		\midrule
		& TOSL+QBIASSR  vs TOSL+SR      & 1.68	        & 0.465     \\
		3D-arm-1K     & TOSL+QBIASSR  vs Q+SR         & 6.20	        & 0.001	    & p$<$0.01 \\
		& TOSL+SR vs Q+SR               & 4.52	        & 0.006	    & p$<$0.01   \\ 
		\midrule
		& TOSL+QBIASSR  vs TOSL+SR     & 2.60	        & 0.165     \\
		3D-arm-4K    & TOSL+QBIASSR  vs Q+SR          & 9.33	        & 0.001	    & p$<$0.01 \\
		& TOSL+SR vs Q+SR               & 6.73	        & 0.001	    & p$<$0.01   \\ 
		\midrule
		& TOSL+QBIASSR  vs TOSL+SR     & 1.95	        & 0.410     \\
		2D-navigation-1K          & TOSL+QBIASSR  vs Q+SR         & 4.07	        & 0.063	    \\
		& TOSL+SR vs Q+SR               & 2.13	        & 0.354	     \\ 
		\bottomrule
	\end{tabular}
	\label{tab:Tukey-VREP}
\end{table*}

\begin{table*}[h!]
	\small\sf\centering
	\caption{ANOVA: Learning techniques on Giraff robot}
	\begin{tabular}{p{.2\textwidth}p{.16\textwidth}p{.12\textwidth}p{.04\textwidth}p{.11\textwidth}p{.055\textwidth}p{.1\textwidth}}
		\toprule
		task            &source             &sum of squares   &df    &mean square   &F         & p-value \\
		\midrule
		&between groups     & 31.73            & 2     & 15.86         & 16.36 & 0.000002\\
		wander-1K      &within groups      & 55.26             & 57    & 0.97 \\  
		&total              & 86.98             & 59 \\
		\bottomrule
	\end{tabular}
	\label{tab:ANOVA-giraff}
\end{table*}

\begin{table*}[h!]
	\small\sf\centering
	\caption{ANOVA: Post-hoc Tukey HSD test on Giraff robot}
	\begin{tabular}{p{.2\textwidth}p{.26\textwidth}p{.14\textwidth}p{.11\textwidth}p{.12\textwidth}}
		\toprule
		task            &pair                           & HSD Q statistic   & HSD p-value    & HSD inference \\
		\midrule
		&TOSL+QBIASSR  vs TOSL+SR       &4.3491	&0.0089151	    &p$<$0.01 \\
		wander-1K    &TOSL+QBIASSR  vs Q+SR          &8.0827	&0.0010053	    &p$<$0.01 \\
		&TOSL+SR vs Q+SR                &3.7336	&0.0283548	    &p$<$0.05  \\          
		\bottomrule
	\end{tabular}
	\label{tab:Tukey-giraff}
\end{table*}

\section*{Funding}
This work has been supported by the Spanish Government through the research project DPI2014-55826-R.

\bibliographystyle{agsm}
\bibliography{references}

\end{document}